\documentclass{bmvc2k}

\title{UDIS: Unsupervised Discovery of Bias in Deep Visual Recognition Models}

\addauthor{Arvindkumar Krishnakumar}{akrishna@gatech.edu}{1}
\addauthor{Viraj Prabhu}{virajp@gatech.edu}{1}
\addauthor{Sruthi Sudhakar}{sruthis@gatech.edu}{1}
\addauthor{Judy Hoffman}{judy@gatech.edu}{1}

\addinstitution{
 Georgia Institute of Technology\\
 Atlanta, GA
}

\runninghead{Krishnakumar et al.}{Unsupervised Discovery of Bias in Deep Models}

\def\etal{\emph{et al}\bmvaOneDot}

\newcommand{\method}{\texttt{UDIS}\xspace}

\usepackage[colorinlistoftodos,prependcaption,textsize=tiny]{todonotes}
\usepackage{float}
\usepackage{paralist}
\usepackage{amsmath}
\usepackage{parskip}
\usepackage{graphicx}
\usepackage{bmpsize}

\DeclareMathOperator*{\argmax}{argmax}
\begin{document}

\maketitle

\begin{abstract}

Deep learning models have been shown to learn spurious correlations from data that sometimes lead to systematic failures for certain subpopulations. Prior work has typically diagnosed this by crowdsourcing annotations for various protected attributes and measuring performance, which is both expensive to acquire and difficult to scale. In this work, we propose \method, an \emph{unsupervised} algorithm for surfacing and analyzing such failure modes. \method identifies subpopulations via hierarchical clustering of dataset embeddings and surfaces systematic failure modes by visualizing low performing clusters along with their gradient-weighted class-activation maps. We show the effectiveness of \method in identifying failure modes in models trained for image classification on the CelebA and MSCOCO datasets. \method is available at \url{https://github.com/akrishna77/bias-discovery}.

\end{abstract}


\section{Introduction}
\label{sec:intro}

Computer vision technology has become increasingly dependent on deep learning models to help make intelligent decisions in high-stakes applications. Such models are often trained on large datasets of images like ImageNet~\cite{russakovsky2015imagenet} and MSCOCO~\cite{lin2015microsoft}, which have been shown to contain implicit biases~\cite{datasetbias, revisetool} that are imbibed and sometimes amplified \cite{shankar2017classification, bolukbasi2016man} by these models. Further, these pretrained models are frequently used as an initialization for other downstream tasks through transfer learning \cite{tan2018survey}. It is thus crucial that in addition to being accurate, models be \emph{fair} and perform equitably across different dataset subpopulations.

However, recent studies have shown several examples where state-of-the-art deep computer vision models learn spurious correlations from their training data which leads to significant performance variance across subpopulations, sometimes across sensitive attributes like race and gender~\cite{burns2019women, propublica, Buolamwini2018GenderSI, zhao2017men, wilson2019predictive}, or even contextual and reporting biases~\cite{eykholt2018robust, devries2019does, rosenfeld2018elephant, beery2018recognition}. Learning such spurious correlations typically leads to poor performance on underrepresented dataset subpopulations and out-of-distribution test data. These models are typically evaluated based on standard performance metrics like test set accuracy, but it is equally important to ensure that the model will perform fairly across different subpopulations when provided with previously unseen data.

Determining whether a trained model is biased is a challenging problem. Prior work has relied on enumerating sensitive attributes (such as race and gender), collecting annotations from domain experts, and measuring performance across these \cite{joo2020gender, aif360-oct-2018, cabrera2019fairvis, brandao2019age, Klare2012FaceRP, krishnapriya2019characterizing}. This process requires considerable manual effort and cost, and is challenging to scale to large datasets.

In this work we present \method, a tool to audit deep learning models for biases before deploying them in the wild. \method discovers subpopulations of the dataset for which the model systematically underperforms, without requiring any protected attribute annotations whatsoever and using only the dataset test split. \method performs hierarchical clustering of dataset embeddings and identifies systematic failure modes by visualizing low performing clusters along with their gradient-weighted class-activation (GradCAM~\cite{Selvaraju_2019}) maps. We show the effectiveness of \method in identifying failure modes in visual recognition models trained on the CelebA and MSCOCO datasets. We make the following contributions:
\begin{itemize}
    \item We present \method, the first unsupervised method for discovering model bias which identifies dataset subpopulations on which the model systematically underperforms, without the need for protected attribute annotations.
    \item We demonstrate the effectiveness of \method at identifying failure modes on the CelebA and MSCOCO datasets.
\end{itemize}
\section{Related Work}
\label{sec:relwork}
While there has been considerable prior work in measuring bias in deep learning models, to the best of our knowledge all of them require apriori knowledge as well as annotations for protected classes across which we desire the model to be unbiased. We summarize these lines of prior work below:

\noindent \textbf{Observational methods}. Torralba and Efros ~\cite{datasetbias, datasetponce} were among the first to stir up the conversation of dataset bias in computer vision, introducing simple measures like cross-dataset generalization and negative set bias to understand how datasets may bias trained models. Recently, Singh~\etal \cite{singh2020dont} proposed the use of statistical information to identify biased categories. They define a category \textit{b} as biased by category \textit{c} if (1) the prediction probability of \textit{b} drops significantly in the absence of \textit{c} and (2) \textit{b} co-occurs frequently with \textit{c}. This requires knowledge of the dataset attributes to determine categories that are biased, along with their co-occurring context category. Other related works tackle the problem of dataset bias by defining algorithms \cite{dwork2011fairness, Khosla_undoingthe, Wang:2020:TFI} and metrics \cite{zhang2018mitigating, pleiss2017fairness, kilbertus2018avoiding, gajane2018formalizing} to establish fairness. In contrast, our method leverages dataset embeddings that can be computed using a forward pass with the model, and is able to identify model biases on the dataset \emph{without} explicit knowledge of protected attributes and their annotations.

\noindent \textbf{Bias detection toolkits}. Most recently, Wang~\etal \cite{revisetool} released an open-source tool that assists in investigating biases within visual datasets, surfacing potential biases along three specific dimensions: object-based, gender-based, and geography-based. Their method however requires datasets to have object, gender, and geography annotations to discover these biases. Two limitations of their method are that i) these annotations may not be easily available, ii) the method would miss failure modes along other dimensions. Further, as they acknowledge, some of their insights are derived from pretrained models and external tools that may themselves contain implicit biases. IBM's AI Fairness 360 \cite{aif360-oct-2018} uses a comprehensive set of metrics, algorithms and mitigation strategies to measure, report and reduce biases in datasets and machine learning models. Similarly, FairML \cite{Adebayo2016FairMLT} is a toolbox that helps audit predictive models by computing the relative significance of the model's inputs. Models are then queried with sample data that emulates real world inputs, and perturbing this data helps determine model fairness. Cabrera~\etal present Fairvis \cite{cabrera2019fairvis}, a visual analytics tool that helps audit fairness in machine learning models by allowing domain experts to investigate subgroups of data, reporting a high-level overview of their performance and suggesting similar subgroups to explore for detecting bias. These methods require full knowledge of the dataset and report bias through well-defined fairness metrics. Our tool works explicitly with visual recognition models and reports bias through underperforming data subpopulations, utilizing visual explanations to understand failure modes. A few methods rely on small image perturbations to determine salient regions of the input image for tasks to establish the presence of bias \cite{chang2019explaining, dabkowski2017real, Fong_2017, goyal2019counterfactual}.

\noindent \textbf{Counterfactual Approaches}. Denton~\etal \cite{denton2020image} and Balakrishnan~\etal \cite{balakrishnan2020causal} present a counterfactual method to identify biases in a smiling attribute classifier. They accomplish this by building a generative model of face images that manipulates specific image characteristics along meaningful factors of variation. They then test how the prediction of the trained classifier changes if a characteristic (deemed irrelevant to the classification task by humans) is altered in a specific targeted manner. They use this technique to identify a causal relationship between features in an image and the classifier output and establish a source of bias. The effectiveness of such methods highly depend on how well the model is able to sufficiently disentangle different image attributes, and ensuring that the newly generated images contain no other significant changes that may affect the outcome of the task. Dash \etal \cite{dash2021evaluating} and Joo and Kärkkäinen \cite{joo2020gender} also propose counterfactual methods to identify bias in visual models. However, they explore bias with respect to specific protected attributes like race and gender. Our method does not require specific sensitive attributes and tries to identify sources of bias of any form that lead to systematic failure modes.
\section{Approach}
\label{sec:approach}

\newcommand{\acc}{\texttt{\textsc{ACC}}}

\begin{figure}[ht]
\begin{center}
\includegraphics[width=0.95\textwidth]{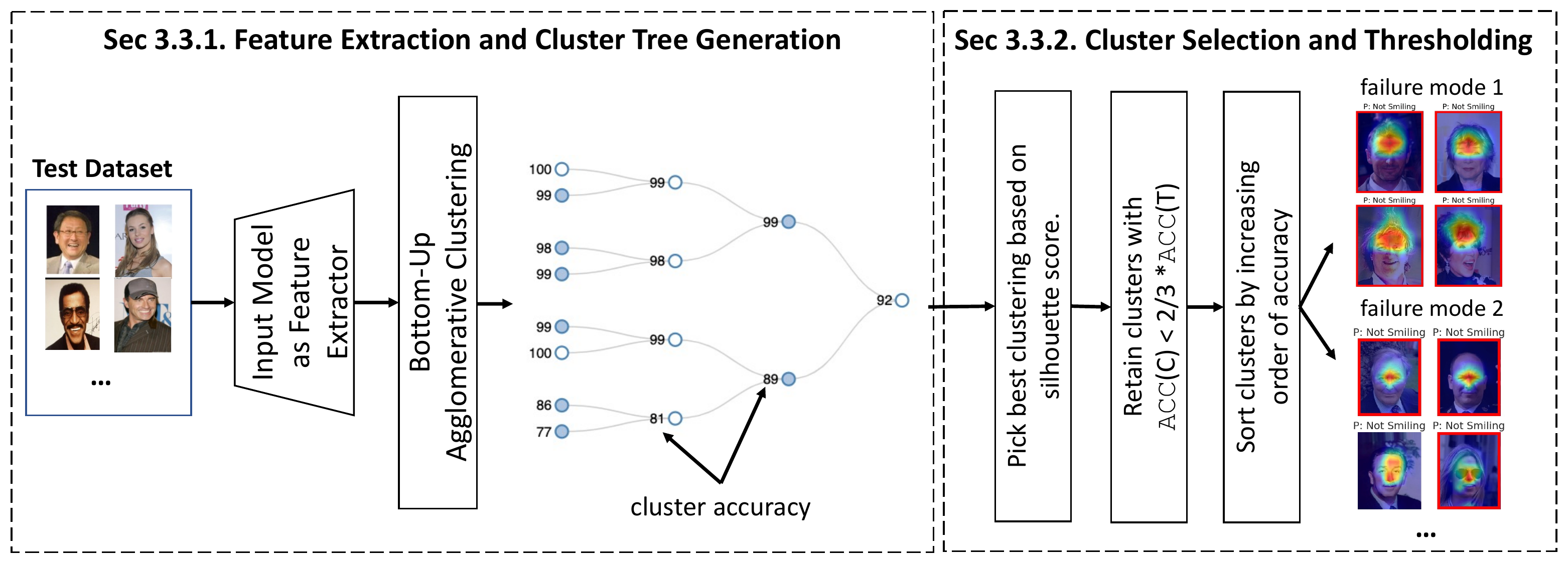}
\end{center}
\vspace{-10pt}
\caption{ We propose \method for unsupervised discovery of biases in a model. \textbf{Left:} The input model is used as a feature extractor for the test dataset. Bottom-up agglomerative clustering is performed on these feature vectors to obtain a binary cluster tree. \textbf{Right:} We use silhouette score as a measure to determine the best clustering from this tree, and filter and sort the clusters based on their accuracies before presenting them to the developer.}
\label{fig:approach}
\end{figure}

We introduce \method for the unsupervised discovery of model biases. We combine a hierarchical clustering technique to discover data subsets deemed similar by the model and use a performance ranking criteria to sort hundreds of clusters and propose to the developer only the few sets most likely to be caused by model bias (see Figure \ref{fig:approach}), eliminating the cost of annotating large-scale data.

Given attribute annotations, prior work \cite{denton2020image} has shown it is possible to learn latent vectors corresponding to semantic concepts, and using these to detect bias via evaluating counterfactual queries. More recent work \cite{peebles2020hessian} has shown that it may be possible to learn such disentangled latent vectors in an unsupervised fashion. But image generation is hard and learning to manipulate one specific attribute at a time is even harder, even in a supervised manner. Further, it is not guaranteed that the learned attributes will be semantic or correspond to features we care about. It is also not clear if this approach will generalize to more complex / smaller datasets. One possible approach is to use off-the-shelf attribute predictors as an alternative to not having attribute annotations, but such models may contain implicit biases themselves. 

Our method utilizes model interpretability, in an effort to find similar sets of images where the model behaves similarly. Since we would like to use this tool mainly for error analysis, we focus on the groups of images for which the model performs poorly. 

\subsection{Notation} 
Let $x$ and $y$ represent the input images and predicted class respectively. Then for a trained convolutional neural network, $ M : x \rightarrow y$, our goal is to identify clusters of similar images that could potentially suggest model biases. For a given input image $x$, the model M generates a $K$-dimensional output (for $K$ classes) for the classification task,
\begin{equation}
    y = \argmax M(x) = \argmax F(h(x)) 
\end{equation} 
where $h(x)$ is the penultimate layer embeddings and $F(.)$ is the final classifier layer.

We define the overall accuracy of the model on the test dataset $T$ as \acc$(T)$ and accuracy of the model on a cluster of images, $C$ as \acc$(C)$. In the multi-label classification setting, when discovering biases with respect to a category \textit{b}, \acc$(T)$ represents the model accuracy with respect to category \textit{b} over the full test dataset $T$.

\subsection{Visual Explanations} 
On retrieving clusters of images, we wish to discover the features of the input image that is responsible for the classification decision. In this regard, we use heatmaps based on GradCAM \cite{Selvaraju_2019} to visualize a mask over the region of the image that the model is focusing on for its classification decision. We compute this mask by computing the gradient of the score for the predicted class $y$, with respect to the feature map activations of the final convolutional layer and global-average-pooling them to obtain importance values for the feature maps. We then apply a ReLU over the weighted linear combination of feature maps and their importances, to obtain a localization heatmap on the region of interest for the class $y$.

\subsection{\method: Unsupervised Discovery of Bias in Deep Visual Recognition Models}
\subsubsection{Feature Extraction and Cluster Tree Generation}

In both the binary and multi-label classification settings, we compute hidden representations $h(x)$ using the penultimate layer of the network (i.e. the layer before the logits layer). For the binary classification setting, we do so for each image in the test dataset, whereas, for multi-label classification, to observe bias with respect to a category \textit{b} we compute $h(x)$ for all the images in the test dataset where the model’s predictions contain the category \textit{b}. 

We then perform hierarchical bottom-up clustering \cite{Inchoate:Ward63} on these hidden representations $h(x)$. We begin with each hidden vector as a singleton cluster, and recursively merge the pair of clusters that leads to the least increase in total within-cluster variance after merging. We use euclidean distance as the metric to compute linkage. This results in a binary tree of image clusters, where leaf nodes represent each image in $T$ as an individual cluster and the root node represents $T$. Parsing the tree from the root, we notice that clustering in this feature space recursively splits clusters into a relatively high accuracy cluster and a relatively low accuracy cluster at every iteration (see Figure \ref{fig:approach}, left). 

\subsubsection{Cluster Selection and Thresholding}

We now present our approach for selecting a set of disjoint and important clusters from our binary cluster tree to present to the developer (see Figure \ref{fig:approach}, right). We begin by exploring the binary tree bottom-up and evaluating the silhouette score \cite{ROUSSEEUW198753} for each cluster at different clustering iterations. Since our method focuses on determining failure modes indicative of model bias, we treat the highest ancestor with 100\% cluster accuracy along any tree branch as a single cluster, while evaluating the silhouette score. 

The silhouette score is a measure of how similar an image is to other images within the same cluster and different from images in other clusters. Our goal is to find a disjoint set of image clusters with the highest silhouette score. Here, the silhouette score for a given clustering refers to the mean silhouette coefficient across all samples. The silhouette coefficient for a single sample is defined using its mean intra-cluster distance $(\mu_\text{intra})$ and its mean nearest-cluster distance $(\mu_\text{near})$ as: 
\begin{equation}
 s = \frac{\mu_\text{near} - \mu_\text{intra}}{\max(\mu_\text{intra}, \mu_\text{near})}    
\end{equation}

The silhouette scores at different clustering iterations form a bitonic sequence, which is strictly increasing, and after the bitonic point, strictly decreasing. This is indicative of poor clustering at the top of the tree where all the images form a single cluster and poor clustering at the bottom of the tree where each image is its own cluster. Thus, the best clustering of images corresponds to the clustering with the bitonic point as its silhouette score. To determine this right set of image clusters optimally, we use a modified binary search. Consider an array of silhouette scores corresponding to every clustering iteration, we check the right subarray if the silhouette score of the array midpoint is part of an increasing subsequence, and the left subarray otherwise. We also impose an additional size constraint on the cluster, to ensure that the smallest cluster contains at least 5 images, and the largest cluster contains no more than 100 images, for the sake of visualization.

For a given clustering $C = \{C_1, C_2, ... C_n\}$, we sort the retrieved clusters in increasing order of their cluster accuracies. Our interest lies in finding failure modes that lead to a large drop in performance. Clusters with small drops in performance compared to \acc$(T)$ tend to be misclassifications or errors and not biases. To surface clusters indicative of bias, we filter the retrieved clusters to obtain $C'$ by dropping the clusters where the cluster accuracy is more than two-thirds of the overall model accuracy on the test set, i.e. 
\begin{equation}
    C' = \left\{C_i~|~ \acc(C_i)~<~\frac{2}{3} * \acc(T) \right\}
\end{equation}  

We experiment with different thresholds to filter the clusters that are potentially indicative of bias. We notice across our different settings that clusters with accuracies below 50\% (for binary problems) are reflective of systematic errors and potentially model bias. To allow for the examination of additional, less obvious or cohesive error types, we return a superset which includes all clusters with accuracy less than 66\% of the overall test accuracy.

For each cluster, $C_i$, we also compute the average feature vector $h_{avg}^{C_i}$ as, 
\begin{equation}
    h_{avg}^{C_i} = \frac{1}{|C_i|} \sum_{x \in C_i} h(x)
\end{equation}
which is used to provide the user with the nearest neighbor cluster with a high accuracy, based on the euclidean distance metric in the feature space. This provides the user with insight on deviant features amongst similar images that may be responsible for failures. If ground truth attribute information is present, the tool also presents the developer with the nearest neighbor cluster with a high accuracy, based on euclidean distance in ground truth \emph{attribute} distribution space (details in supplementary material). \\

\vspace{-.5cm}
\section{Experiments}
\label{sec:experiments}

\subsection{Overview}
\label{sec:experiments-overview}

We show the results of our method for three settings -- two single attribute prediction tasks on the CelebA~\cite{liu2015faceattributes} dataset and multilabel classification on the MS COCO~\cite{lin2015microsoft} dataset. 
\begin{compactenum}
    \item \textbf{Smiling prediction on CelebA.} We train a Resnet50~\cite{he2015deep} backbone (initialized with ImageNet weights) on the CelebA dataset to predict if a person is Smiling/Not Smiling. The trained model has an accuracy of 92\% on test data.
    
    \item \textbf{Smiling prediction on biased CelebA.} In this setting, we \emph{intentionally} induce bias in the dataset towards the ``Black Hair'' attribute. We do this by manually subsampling the training dataset to increase the proportion of images containing the ``Black Hair'' attribute that are labeled as  ``Smiling''. Conversely, we increase the proportion of images \emph{not} having the ``Black Hair'' attribute that are labeled as ``Not Smiling''. 
    We ensure that the despite the induced bias, we have a model that performs well on test data (93\% accuracy).
    \item \textbf{Multilabel classification on MS COCO.} As \method is model agnostic, we also include a multi-label 80-way classification task. We use an open-source DenseNet~\cite{huang2017densely, huang2019convolutional} classifier trained on the MSCOCO dataset from Wang~\etal \cite{wang2019elastic}, that uses a binary cross-entropy loss to predict multiple labels for an input image.
\end{compactenum}

\subsection{Implementation details}

The ResNet50 models are trained with PyTorch~\cite{paszke2019pytorch} on 8 NVIDIA RTX 2080 GPUs with the SGD optimizer, batch size $64$, weight decay $1\times10^{-4}$, learning rate $5\times10^{-4}$, momentum $0.99$ and a dropout of $0.3$. The model accuracy in both the settings is comparable to that presented in Denton~\etal \cite{denton2020image}.

For \method, we cluster the 2048-dimensional average pooled outputs of the `layer4' module of the ResNet50 model. For the DenseNet model, we use the 1920-dimensional output of the final BatchNorm layer (`norm5') at the end of the dense blocks.
\begin{figure}[h!]
    \centering
    \begin{subfigure}[]{0.45\linewidth}
        \centering
        \includegraphics[width=\linewidth]{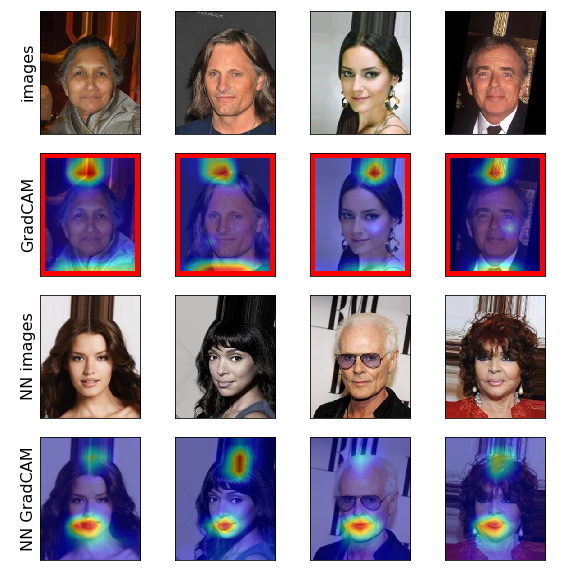}
        \caption{}
        \label{fig:2a}
    \end{subfigure}
    \qquad
    \begin{subfigure}[]{0.45\linewidth}  
        \centering
        \includegraphics[width=\linewidth]{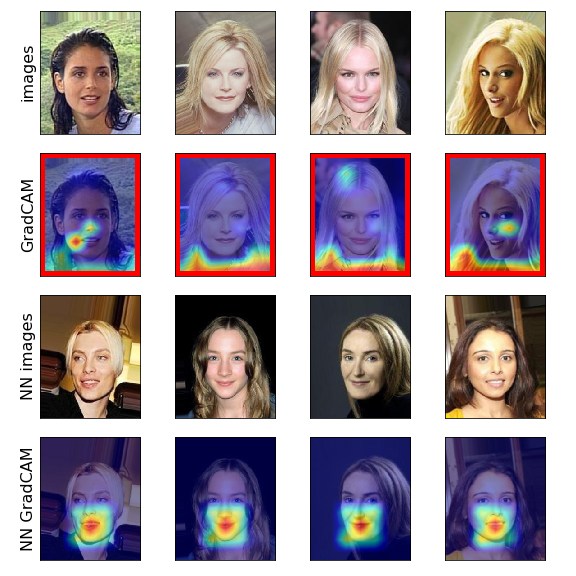}
        \caption{}
        \label{fig:2b}    
    \end{subfigure}
    \vspace*{3mm}
    \caption{\textbf{Bias Discovery on CelebA:} Example visualizations of discovered biases (\textit{top row}) and their GradCAM heatmaps (\textit{second row}) using the model trained to predict smiling on the original CelebA dataset. The bottom two rows represent the nearest neighbor cluster with similar attributes but high accuracy, to similar samples where the model performs well. A red frame indicates an incorrect classification.}
    \label{fig:original}
\end{figure}

\begin{figure}[ht] 
    \centering
    \begin{subfigure}[]{0.45\linewidth}
        \centering
        \includegraphics[width=\linewidth]{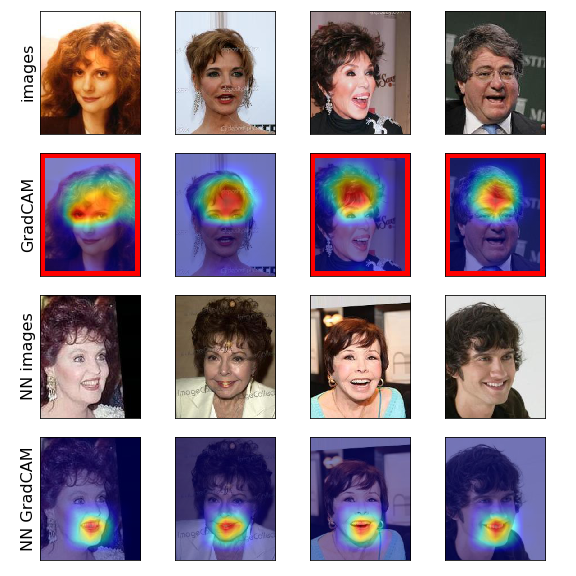}
        \caption{}
        \label{fig:3a}
    \end{subfigure}
    \quad
    \begin{subfigure}[]{0.45\linewidth}  
        \centering
        \includegraphics[width=\linewidth]{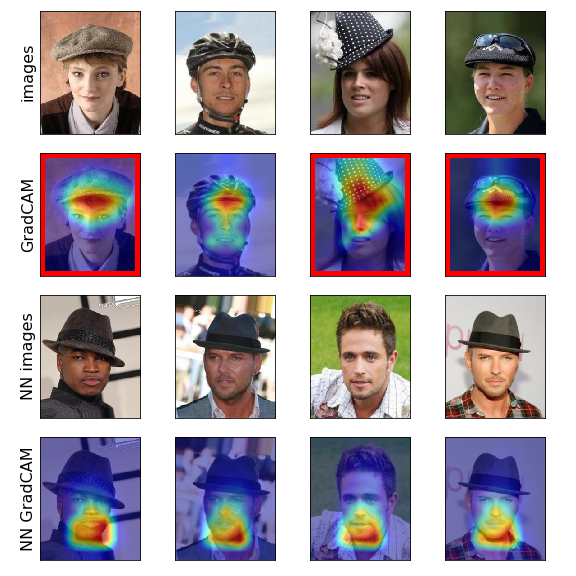}
        \caption{}
        \label{fig:3b}    
    \end{subfigure} 
    \vspace*{3mm}
    \caption{\textbf{Bias Discovery on Black Hair Biased CelebA:} Example visualizations of discovered biases and their GradCAM heatmaps using the model trained on the the biased CelebA dataset for smiling prediction. The bottom two rows represent the nearest neighbor cluster with similar attributes but high accuracy, to showcase the bias in model output. A red frame indicates an incorrect classification.}
    \label{fig:biasedbh}
\end{figure}

\vspace{-.5cm}
\subsection{Biases Discovered by \method}

\begin{figure}[h!]
    \centering
    \begin{subfigure}[]{0.4\linewidth}
        \centering
        \includegraphics[width=\linewidth]{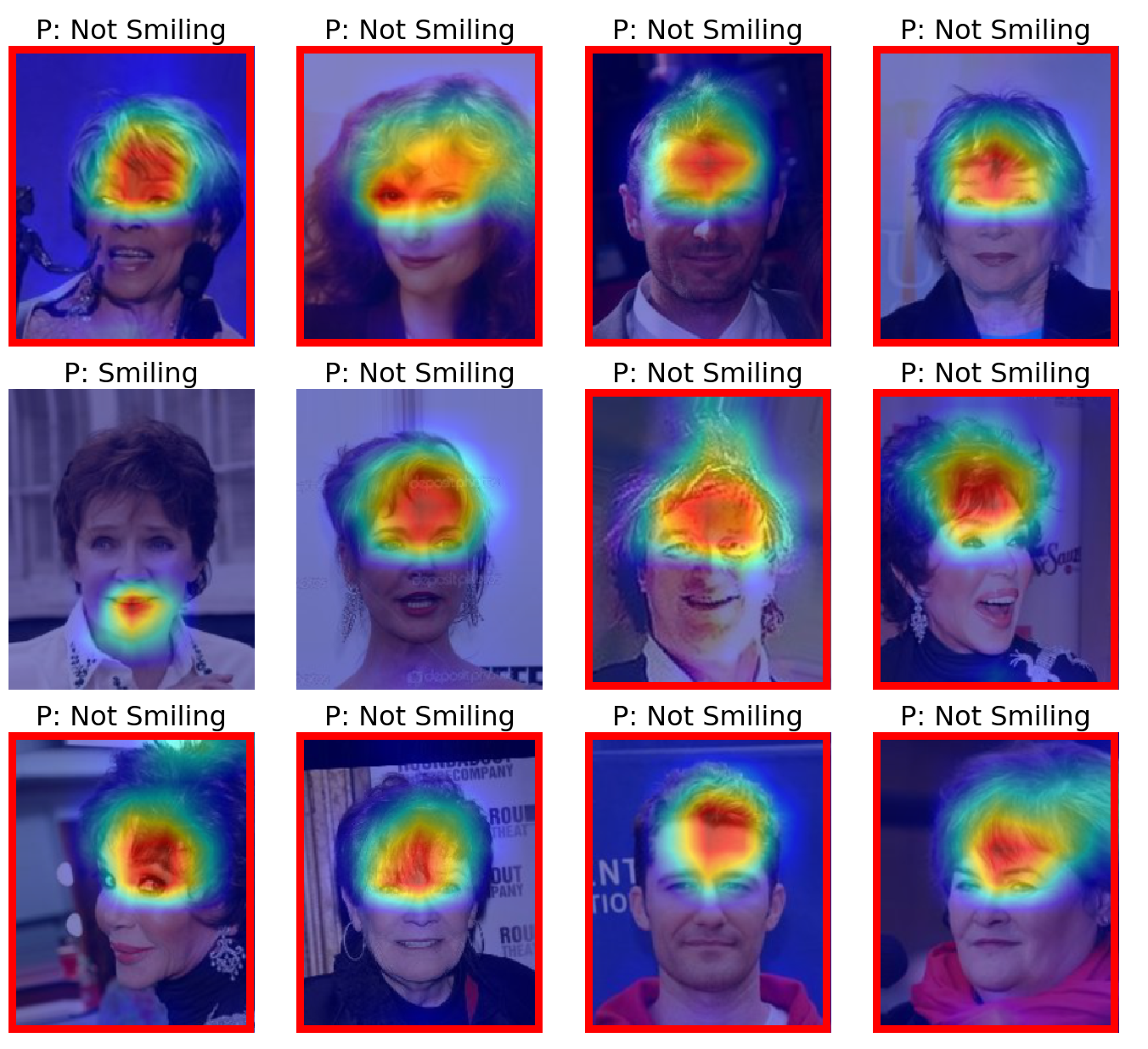}
        \caption{}
        \label{fig:5a}
    \end{subfigure}
    \quad
    \begin{subfigure}[]{0.4\linewidth}  
        \centering
        \includegraphics[width=\linewidth]{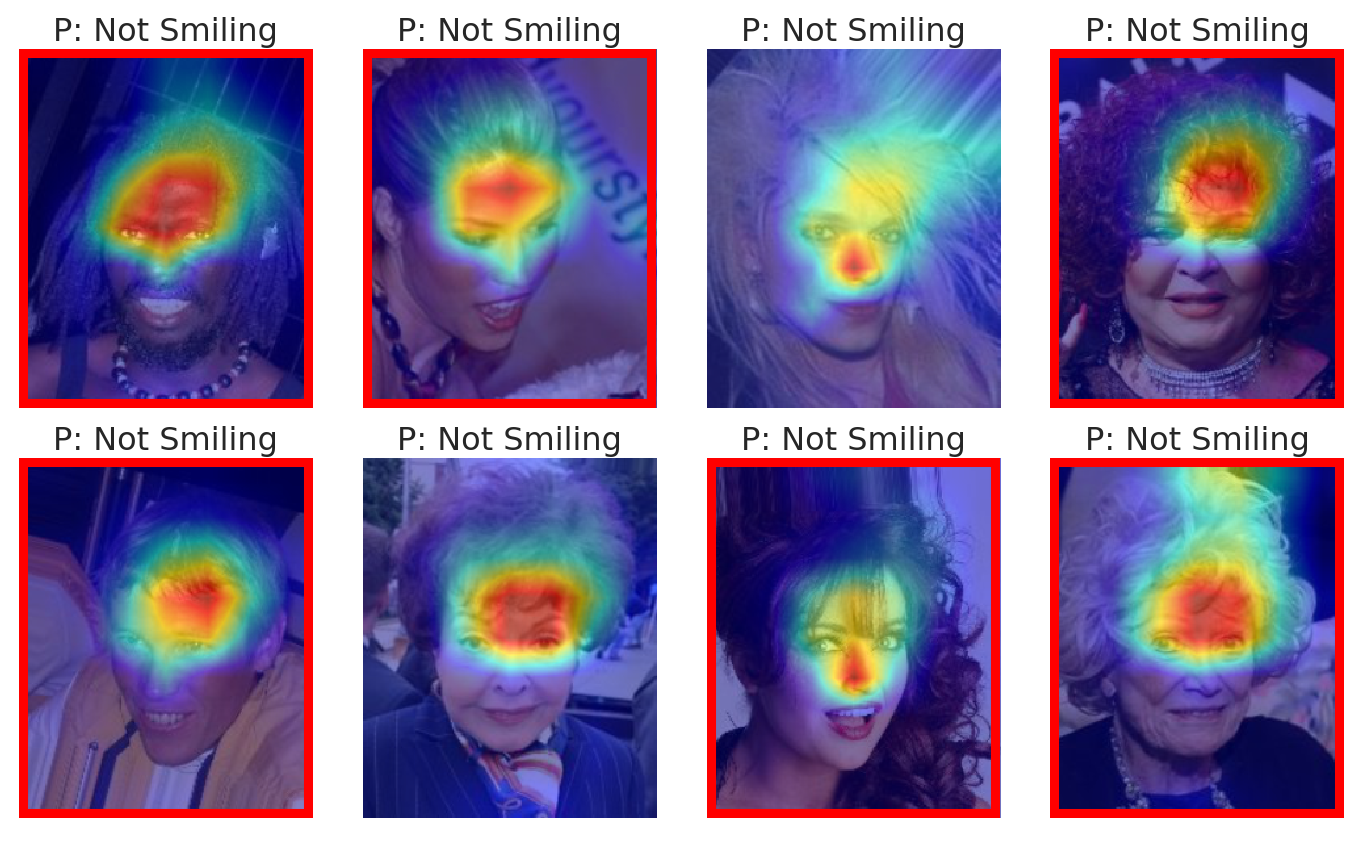}
        \caption{}
        \label{fig:5b}    
    \end{subfigure}
    \vspace*{3mm}
    \caption{\textbf{Subpopulations discovered:} GradCAM heatmaps for the top 2 clusters discovered using \method on the model trained on the \emph{biased} CelebA dataset for smiling prediction (more clusters in supplementary material). 
    }
    \label{fig:clusters}
\end{figure}

In Figures~\ref{fig:original}-~\ref{fig:clusters}, we present some of the biases discovered by \method across settings.
In Figures \ref{fig:original} and \ref{fig:biasedbh}, the top half shows images from the discovered cluster and their visual explanations for the classification decision using GradCAM\cite{Selvaraju_2019}. The bottom half presents the nearest neighbor cluster with high accuracy and their corresponding visual explanations. 

\noindent \textbf{Smiling prediction on CelebA.} In Figure \ref{fig:original}, we present the discovered clusters using the model trained on the original CelebA dataset. We notice from the top half of Figure \ref{fig:2a} that the model is basing its decision for predicting if the person in the image is Smiling, on the artifact above the persons head (the complete cluster containing these images is included in the supplement). Clearly, the unsupervised nature of our method allows for the discovery of spurious correlations or artifacts in the image that may not correspond to an human-identifiable visual attribute (see Figure \ref{fig:2a}).

This pattern is observed in a considerable number of images, and is frequently responsible for incorrect classifications. For instance, the bottom half of Figure~\ref{fig:2a} displays a cluster containing a similar artifact, but the model is able to focus on the right region of the image, i.e. the mouth. Domain experts may be better equipped to infer subtle differences between the images that lead to erroneous classifications. In Figure \ref{fig:2b}, we present sample images from another cluster where the model focuses on the region surrounding the collarbone to make its classification decision. 
Neither of these regions are relevant to the task of smiling prediction itself, and thus can be considered to be indicative of model bias.

\noindent \textbf{Smiling prediction on Biased CelebA.} Since \method is able to discover spurious correlations that the model has learned, with our next experiment we hope to uncover a specific bias that we intentionally induce into the dataset. Under the model trained on the CelebA dataset with a bias toward the ``Black Hair'' attribute, the top clusters discovered correspond to visual explanations in the region of the image surrounding the hair (see Figure \ref{fig:biasedbh}).

Sample images from the top cluster (full cluster in supplementary material) are shown in Figure \ref{fig:3a}, along with similar images from a higher accuracy cluster and their corresponding visual explanations. The tool also finds clusters (see supplementary material) where it discovers a bias learned against people wearing hats, as shown in Figure \ref{fig:3b}. This is still consistent with our experimental setting, as ``Wearing Hat'', is likely accompanied by the ``Black Hair'' attribute being False. Examples of the subpopulations discovered by \method in this experimental setting can be seen in Figure \ref{fig:clusters}.

\begin{figure}[h!]
    \centering
    \begin{subfigure}[]{0.4\linewidth}
        \centering
        \includegraphics[width=\linewidth]{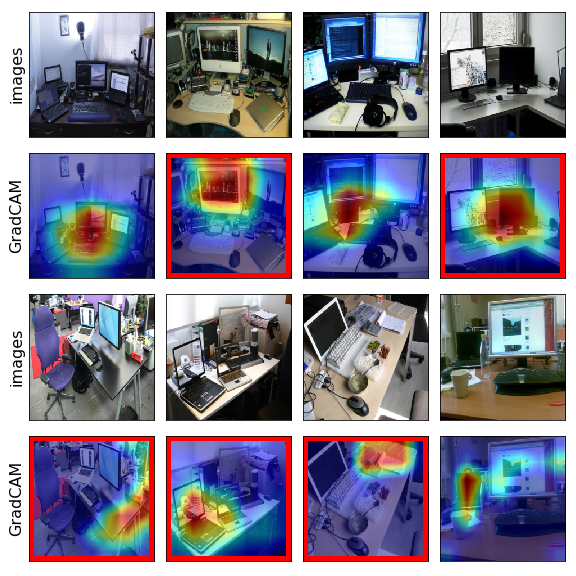}
        \caption{Cluster discovered under the predicted category \textit{cup}, along with GradCAM heatmap for the category \textit{cup}}
        \label{fig:4a}
    \end{subfigure}
    \quad
    \begin{subfigure}[]{0.4\linewidth}  
        \centering
        \includegraphics[width=\linewidth]{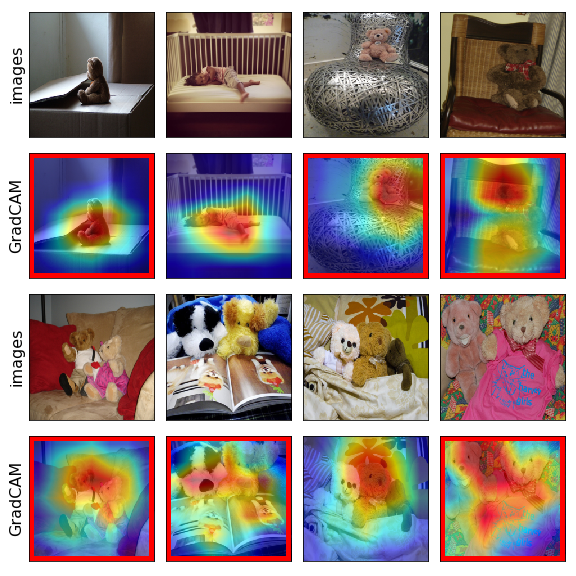}
        \caption{Cluster discovered under the predicted category \textit{bed}, along with GradCAM heatmap for the category \textit{bed}}
        \label{fig:4b}    
    \end{subfigure}
    \vspace*{3mm}
    \caption{\textbf{Bias Discovery on COCO:} Example visualizations of discovered biases (\textit{top row}) and their GradCAM heatmaps (\textit{second row}) using the model trained on the COCO dataset for the multilabel classification task. A red frame indicates an incorrect classification as the predicted category.}
    \label{fig:coco}
\end{figure}

\noindent \textbf{Multilabel classification on MS COCO.} We present the results of biases discovered against 2 predicted categories: \textit{cup} and \textit{bed} (see Figure \ref{fig:coco}). When exploring the validation dataset of MSCOCO for biases against the category \textit{cup}, sample images from the top cluster returned by our method can be seen in Figure~\ref{fig:4a}. The model predicts all the images shown as containing the \textit{cup} class. We notice that all the images in the cluster (see supplementary for full cluster) contain a screen and it is likely that the model associates the frequent occurrence of a cup next to a screen in the training dataset, to overpredict ``cup'' whenever it sees a screen. 
In Figure \ref{fig:4b}, we show example images of biases discovered for the category \textit{bed}. The model seems to have learnt a spurious correlation between \textit{bed} and \textit{teddy bear} from the training data, labelling a number of instances where the \textit{teddy bear} occurs without the \textit{bed}, as a \textit{bed}. 
\section{Conclusion}
\label{sec:conclusion}

In conclusion, we present \method, an unsupervised method which is able to automatically discover subpopulations of visual datasets where the model systematically underperforms. We demonstrate using visual explanations that these subpopulations contain potential biases, and leave it to model developers to investigate the cause of such biases, evaluate their importance and take further action. 

We note some important limitations of our method. \method exclusively focuses on discovering bias in the form of failure modes, where \emph{bias} is defined as any spurious (\emph{i.e.} irrelevant to the task at hand) correlation learnt by the model that \emph{leads to a drop in test accuracy}. We acknowledge that this is only a subset of all possible encoded bias, as in some cases spurious correlations may potentially also improve model performance. Further, the model may also have failure modes due to optimization or generalization error that do not represent model bias. Finally, it remains an open question to determine how frequently the bias discovered by \method correlates with known cultural biases. 

In summary, we emphasize that our method does not detect all possible source of model bias, that each failure mode discovered may not always correspond to a model bias, and further even the ones that do, may not represent an ``interpretable'' bias against a protected attribute such as race or gender.
Extending our method to discover other kinds of bias, including those that lead to improvements in performance is a promising line of future work.

\noindent\textbf{Acknowledgments:} This work was funded in part by Cisco Inc. 
\bibliography{egbib}
\newpage
\appendix

\section{Appendix}
\label{appendix:a}
In this section, we showcase the results of our method and present several failure modes discovered by \method. We use our method to discover potential biases under different settings described in Section \ref{sec:experiments-overview}.
\newpage
\subsection{Results on CelebA - Biased Black Hair Setting}
\label{appendix:bbh}

\begin{figure}[H] 
    \centering
    \includegraphics[width=\textwidth,height=0.9\textheight,keepaspectratio]{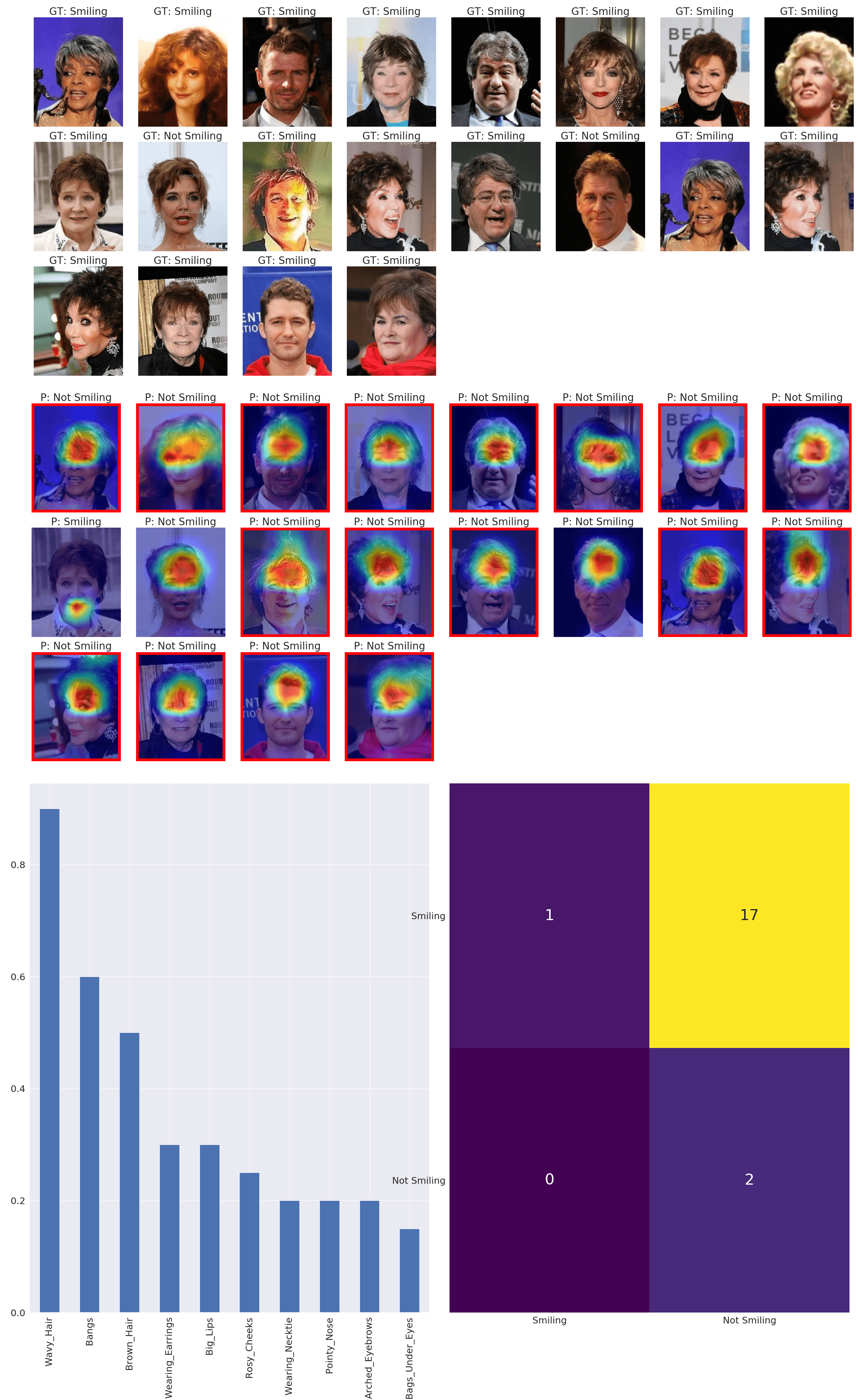}
    \caption{Cluster \#1}
    \label{fig:bbh-1}
\end{figure}

\begin{figure}[H] 
    \centering
    \includegraphics[width=\textwidth,height=0.9\textheight,keepaspectratio]{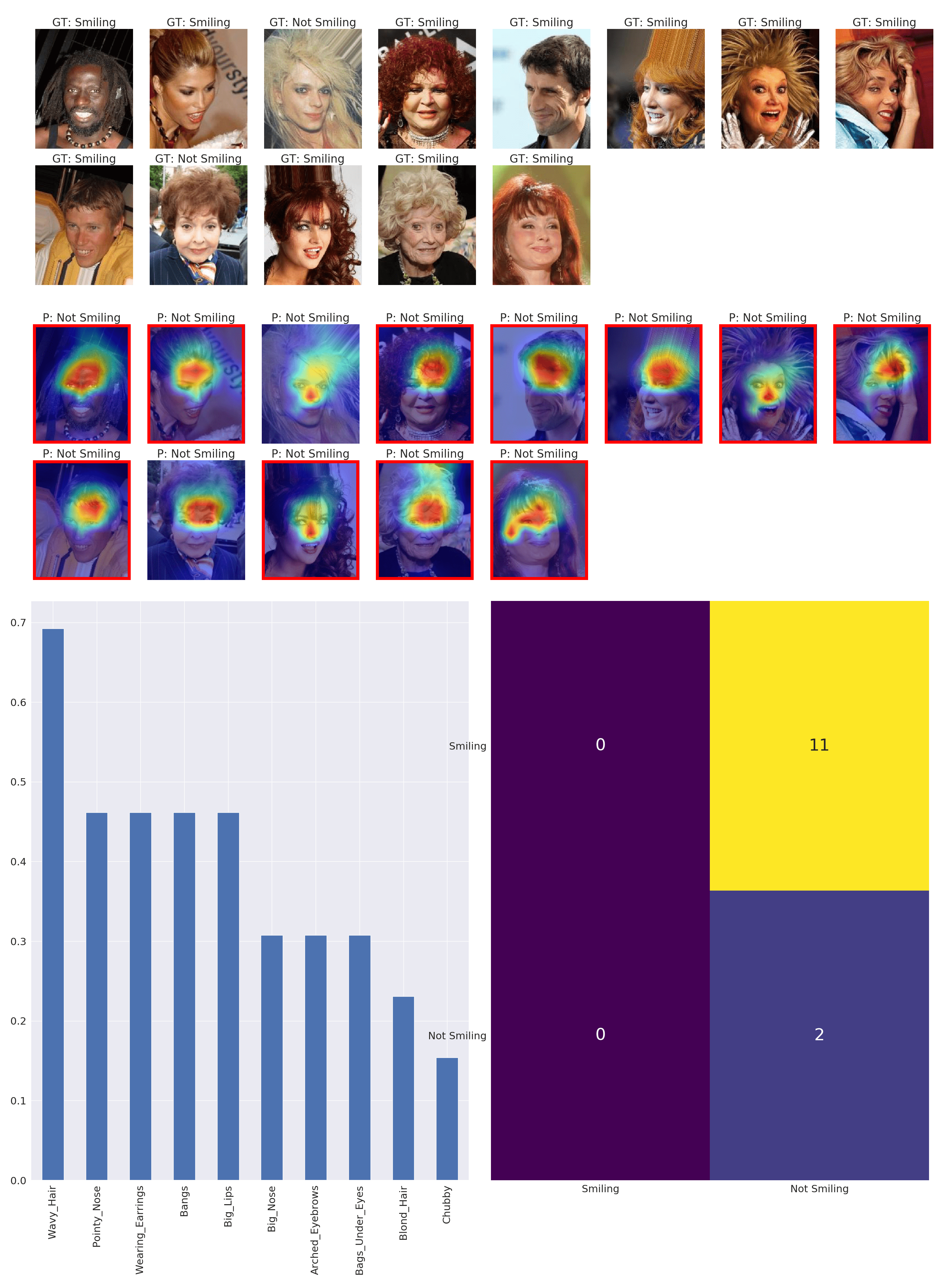}
    \caption{Cluster \#2}
    \label{fig:bbh-2}
\end{figure}

\begin{figure}[H] 
    \centering
    \includegraphics[width=\textwidth,height=0.9\textheight,keepaspectratio]{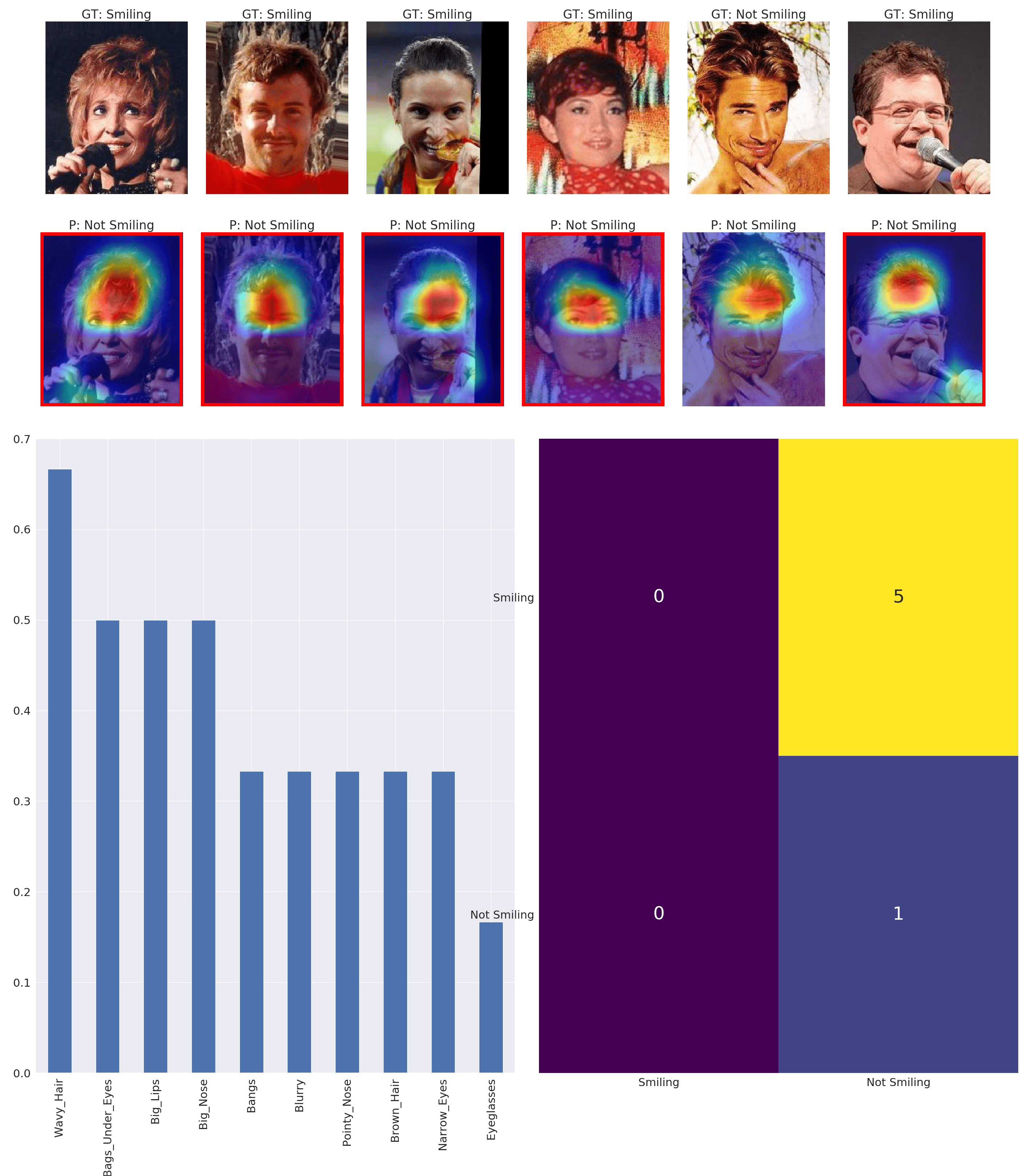}
    \caption{Cluster \#3}
    \label{fig:bbh-3}
\end{figure}

\begin{figure}[H] 
    \centering
    \includegraphics[width=\textwidth,height=0.9\textheight,keepaspectratio]{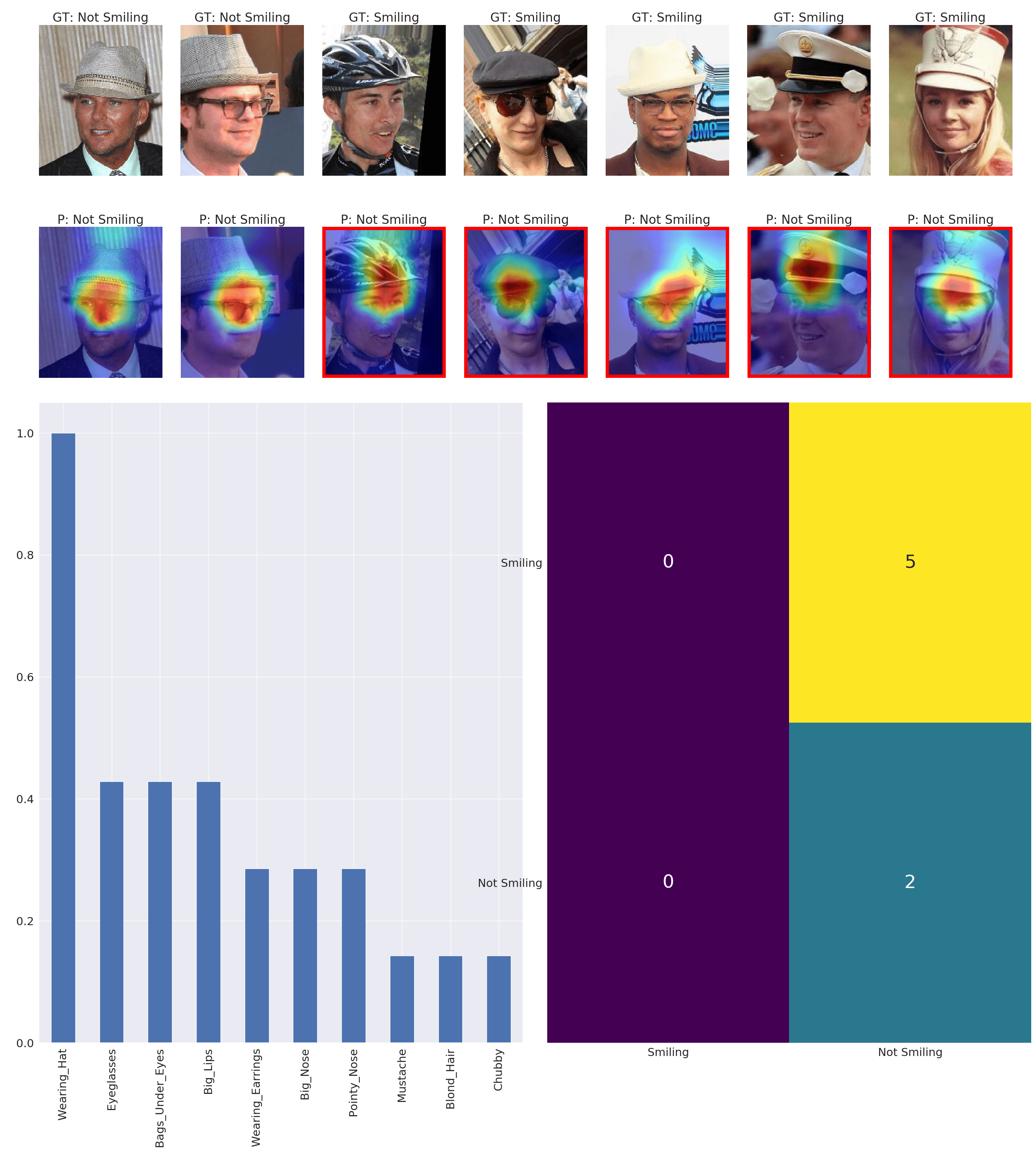}
    \caption{Cluster \#4}
    \label{fig:bbh-4}
\end{figure}

\begin{figure}[H] 
    \centering
    \includegraphics[width=\textwidth,height=0.9\textheight,keepaspectratio]{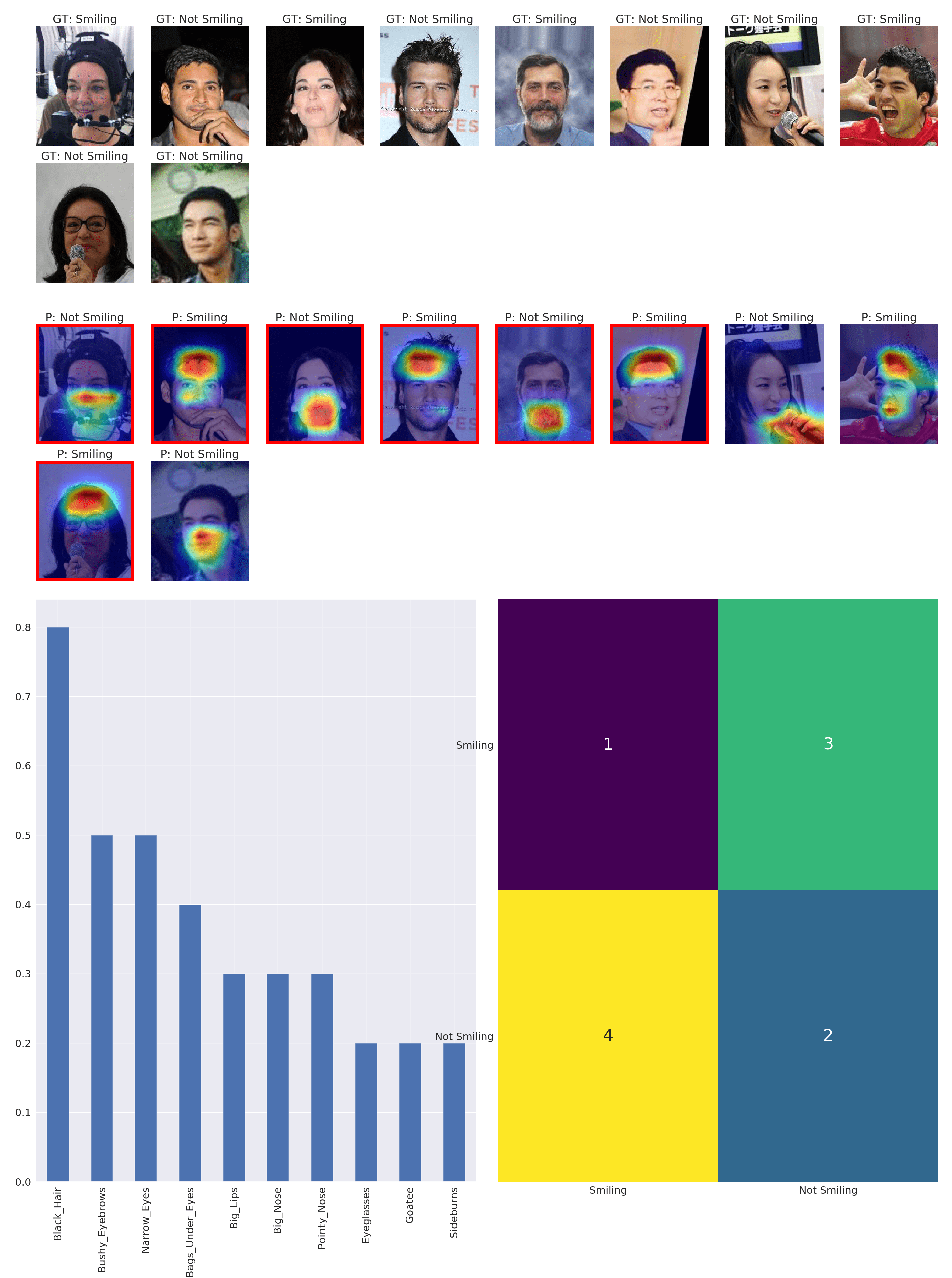}
    \caption{Cluster \#5}
    \label{fig:bbh-5}
\end{figure}

\begin{figure}[H] 
    \centering
    \includegraphics[width=\textwidth,height=0.9\textheight,keepaspectratio]{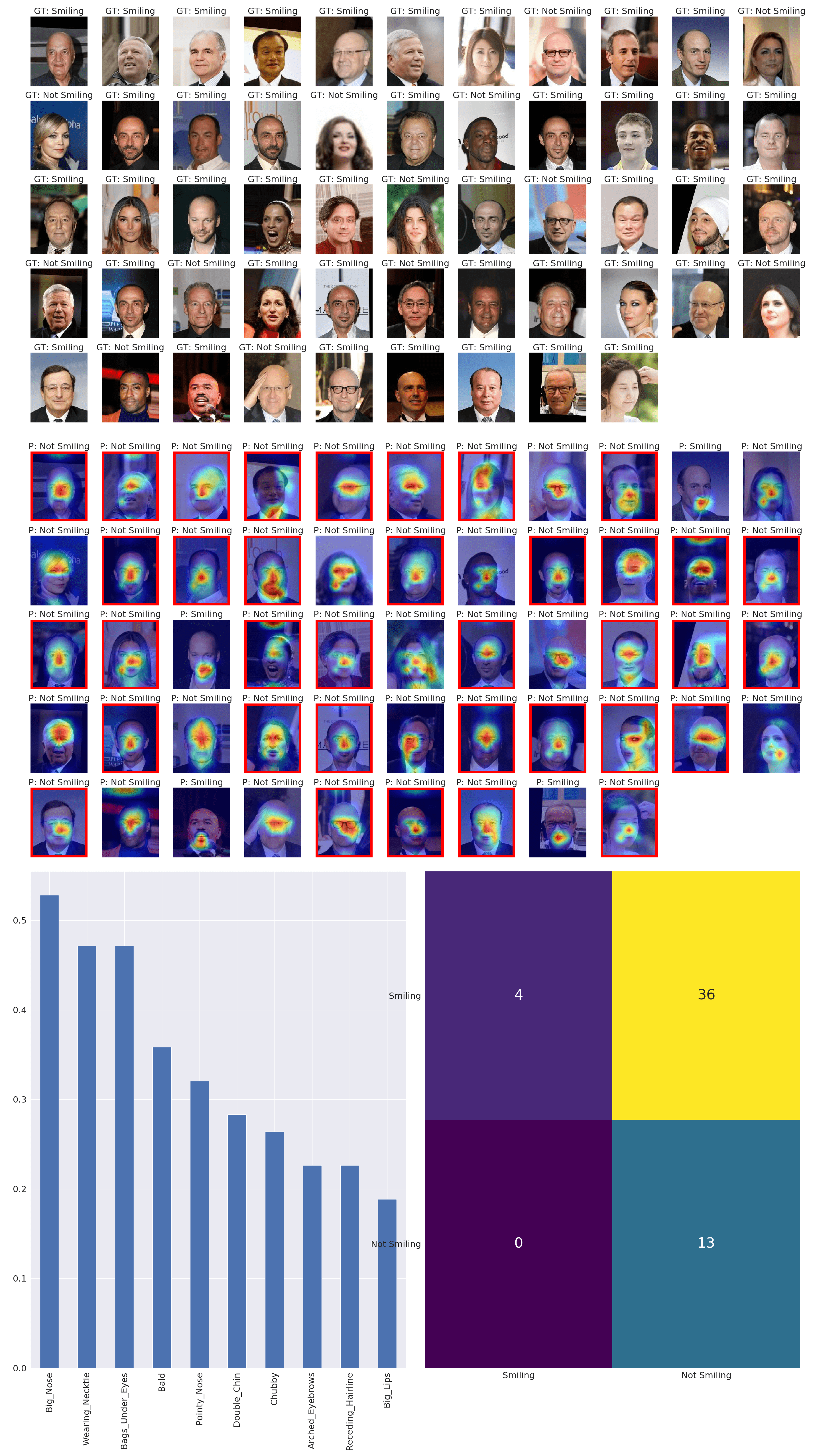}
    \caption{Cluster \#6}
    \label{fig:bbh-6}
\end{figure}

\begin{figure}[H] 
    \centering
    \includegraphics[width=\textwidth,height=0.9\textheight,keepaspectratio]{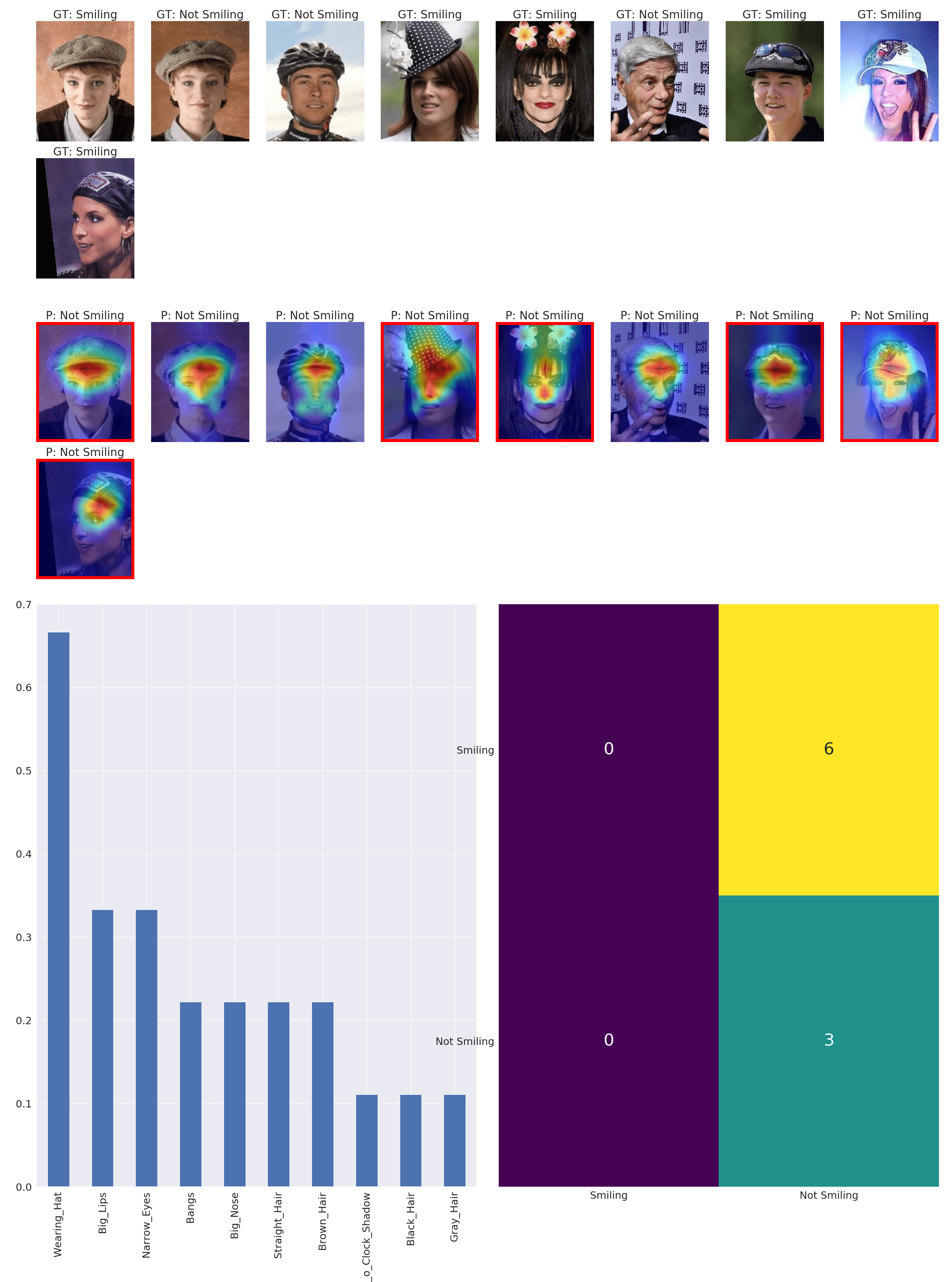}
    \caption{Cluster \#7}
    \label{fig:bbh-7}
\end{figure}

\begin{figure}[H] 
    \centering
    \includegraphics[width=\textwidth,height=0.9\textheight,keepaspectratio]{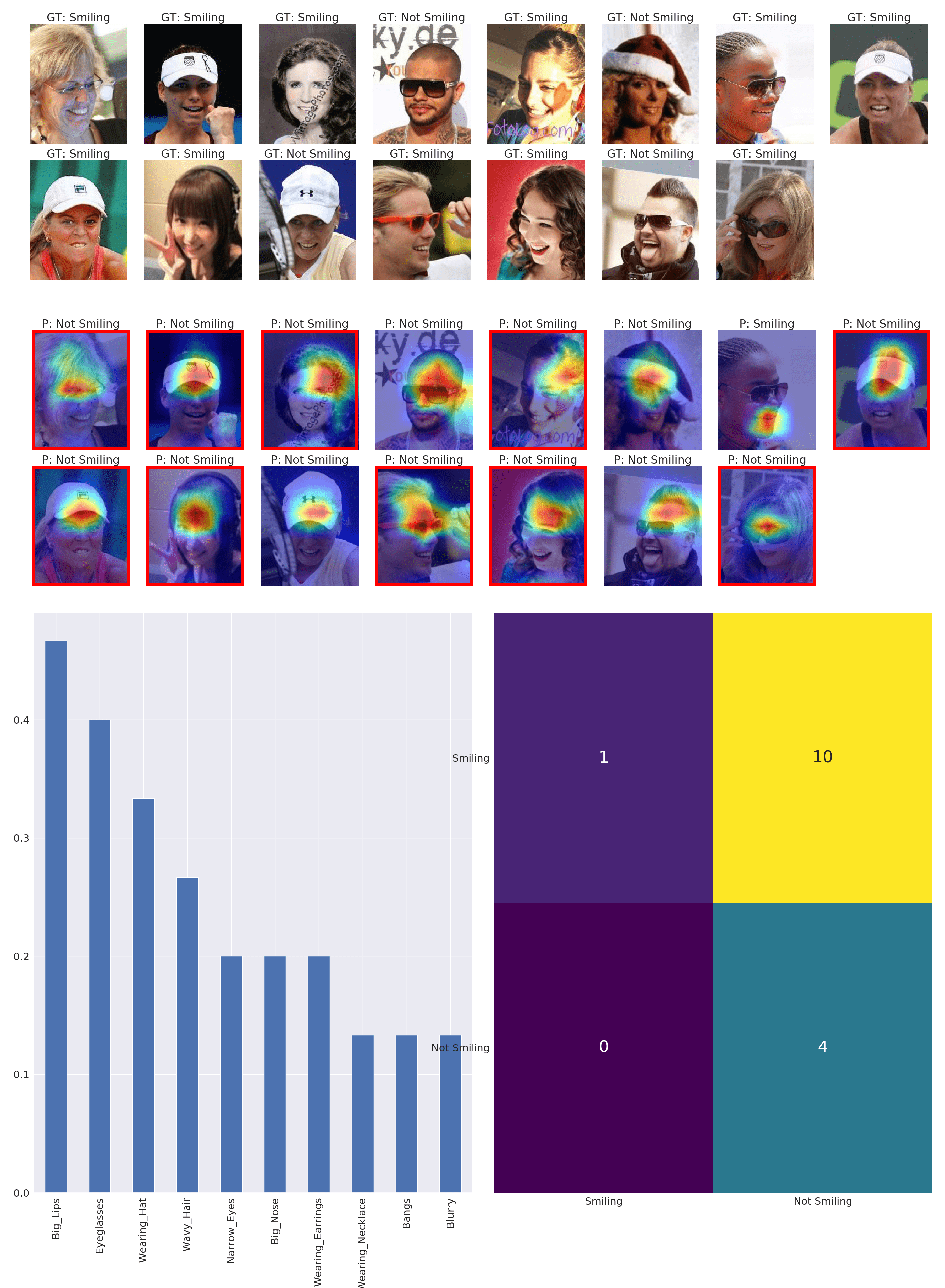}
    \caption{Cluster \#8}
    \label{fig:bbh-8}
\end{figure}

\begin{figure}[H] 
    \centering
    \includegraphics[width=\textwidth,height=0.9\textheight,keepaspectratio]{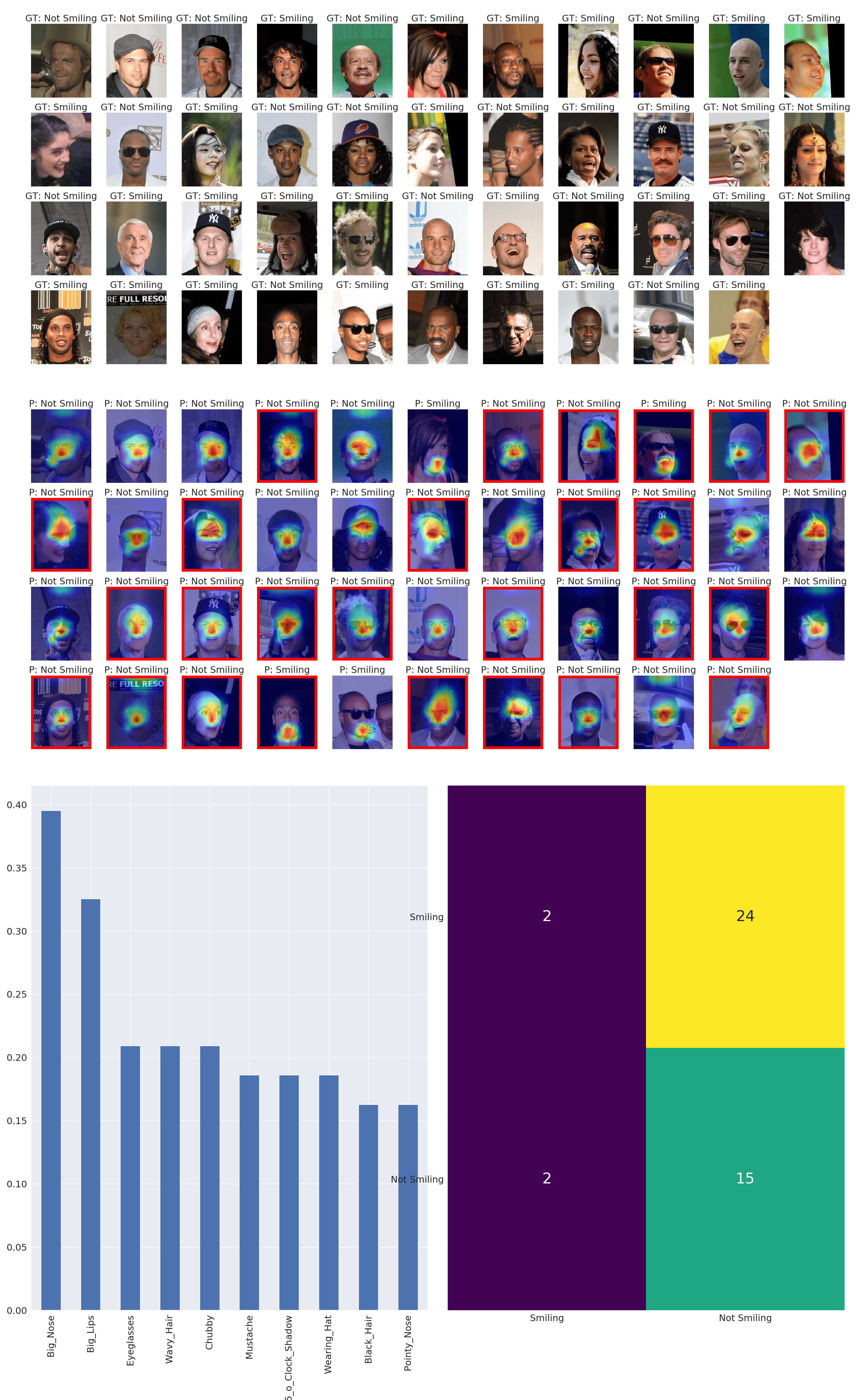}
    \caption{Cluster \#9}
    \label{fig:bbh-9}
\end{figure}

\begin{figure}[H] 
    \centering
    \includegraphics[width=\textwidth,height=0.9\textheight,keepaspectratio]{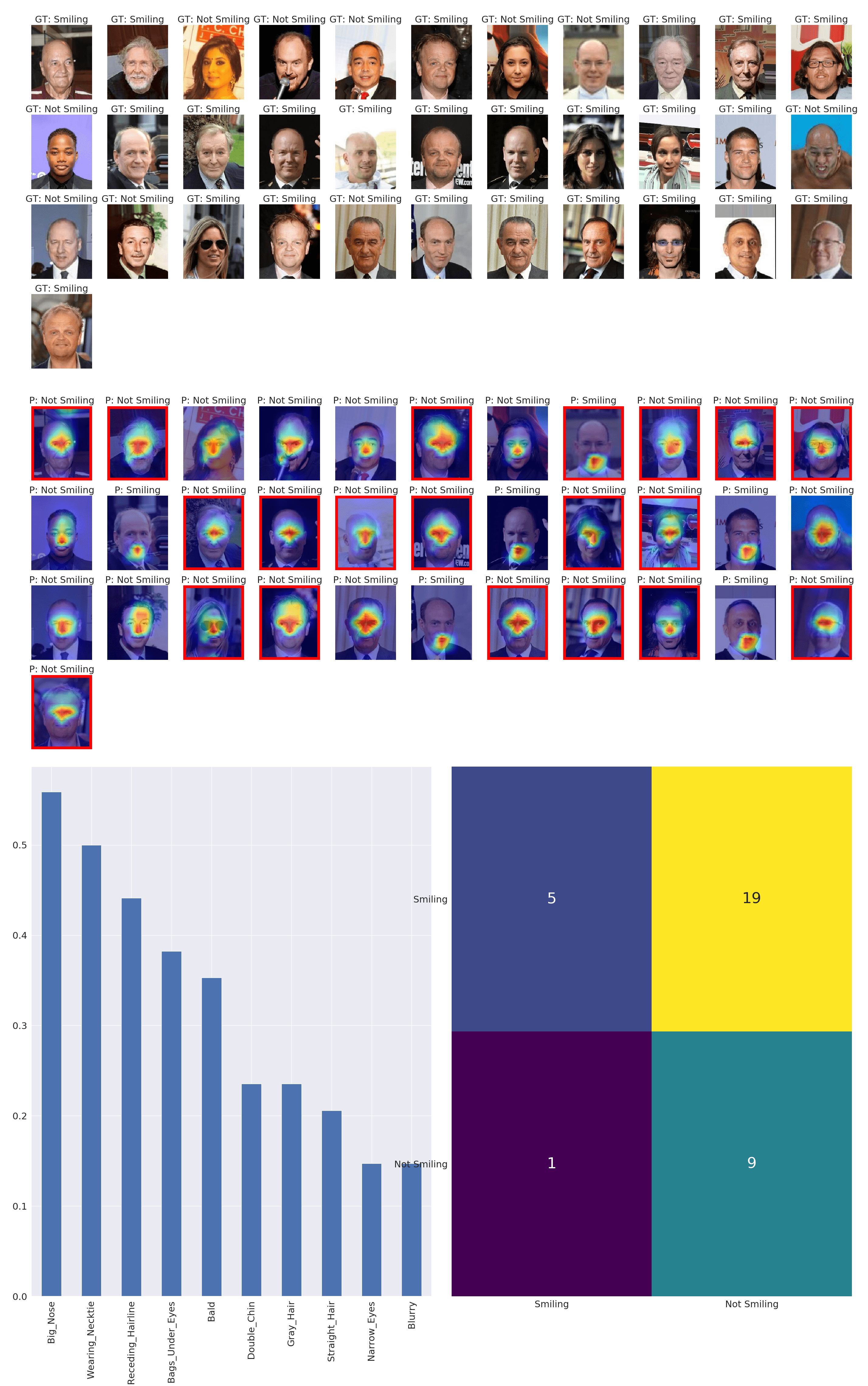}
    \caption{Cluster \#10}
    \label{fig:bbh-10}
\end{figure}

\subsection{Results on CelebA}
\label{appendix:standard}

\begin{figure}[H] 
    \centering
    \includegraphics[width=\textwidth,height=0.9\textheight,keepaspectratio]{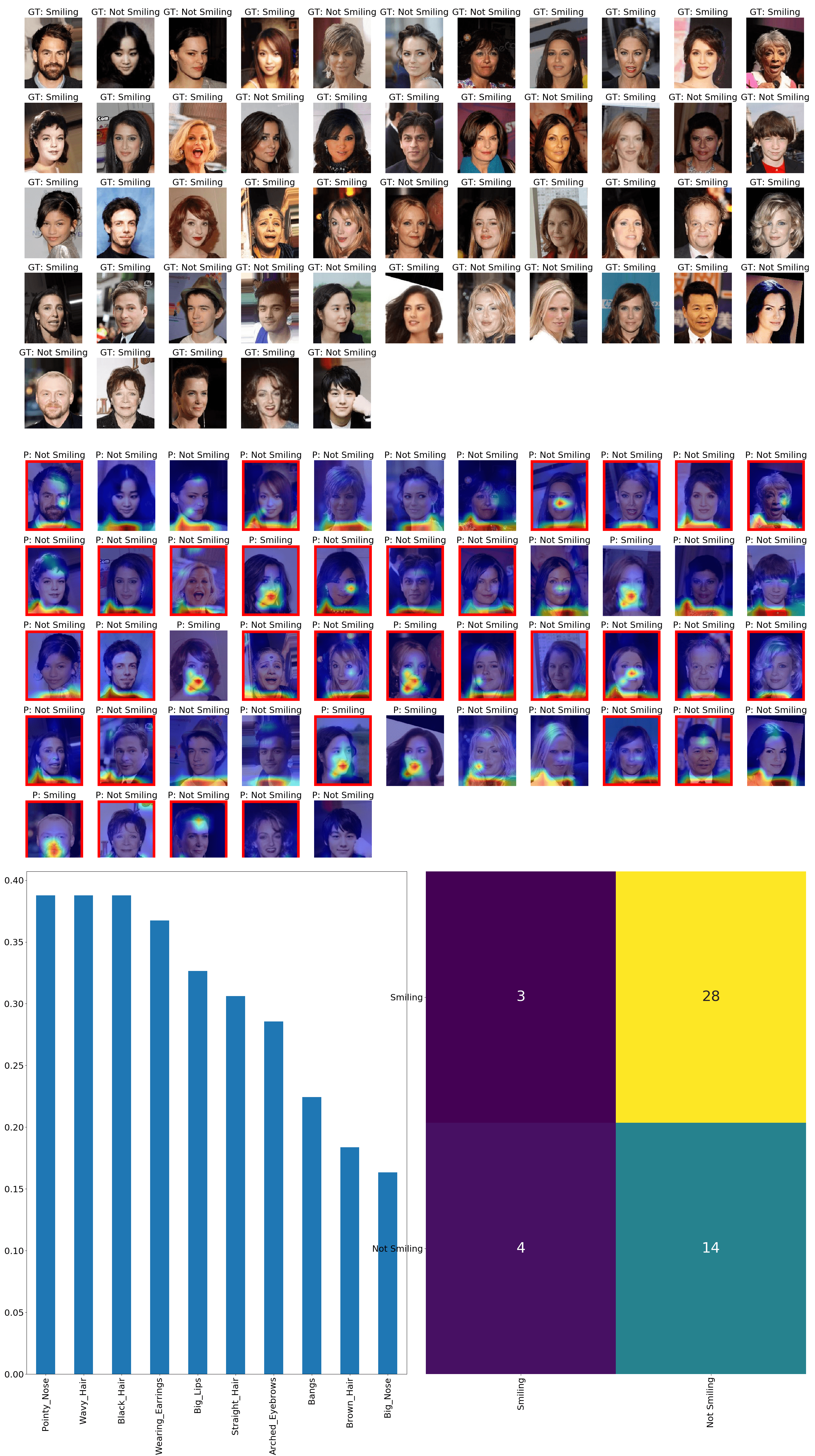}
    \caption{Cluster \#1}
    \label{fig:og-1}
\end{figure}

\begin{figure}[H] 
    \centering
    \includegraphics[width=\textwidth,height=0.9\textheight,keepaspectratio]{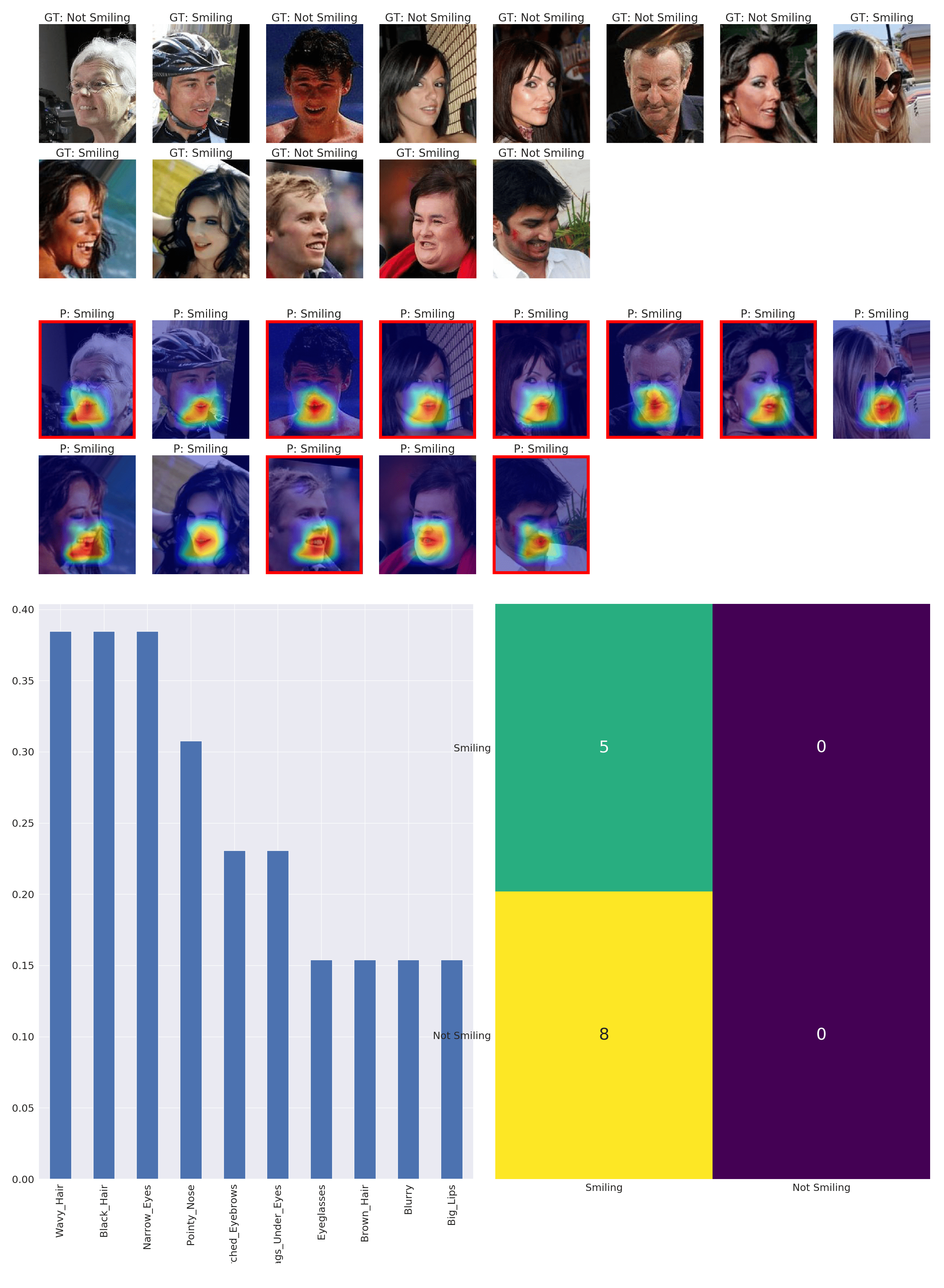}
    \caption{Cluster \#2}
    \label{fig:og-2}
\end{figure}

\begin{figure}[H] 
    \centering
    \includegraphics[width=\textwidth,height=0.9\textheight,keepaspectratio]{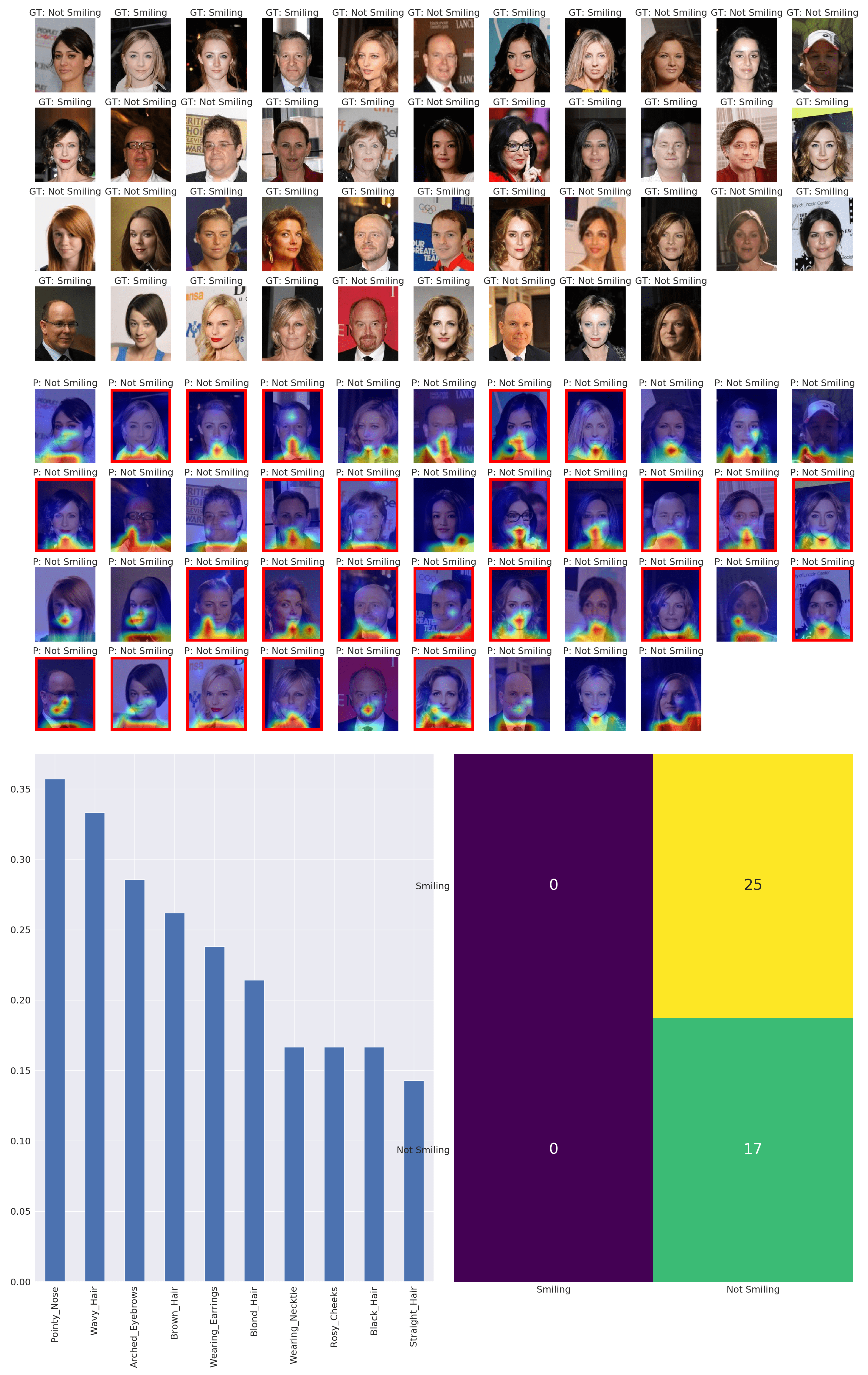}
    \caption{Cluster \#3}
    \label{fig:og-3}
\end{figure}

\begin{figure}[H] 
    \centering
    \includegraphics[width=\textwidth,height=0.9\textheight,keepaspectratio]{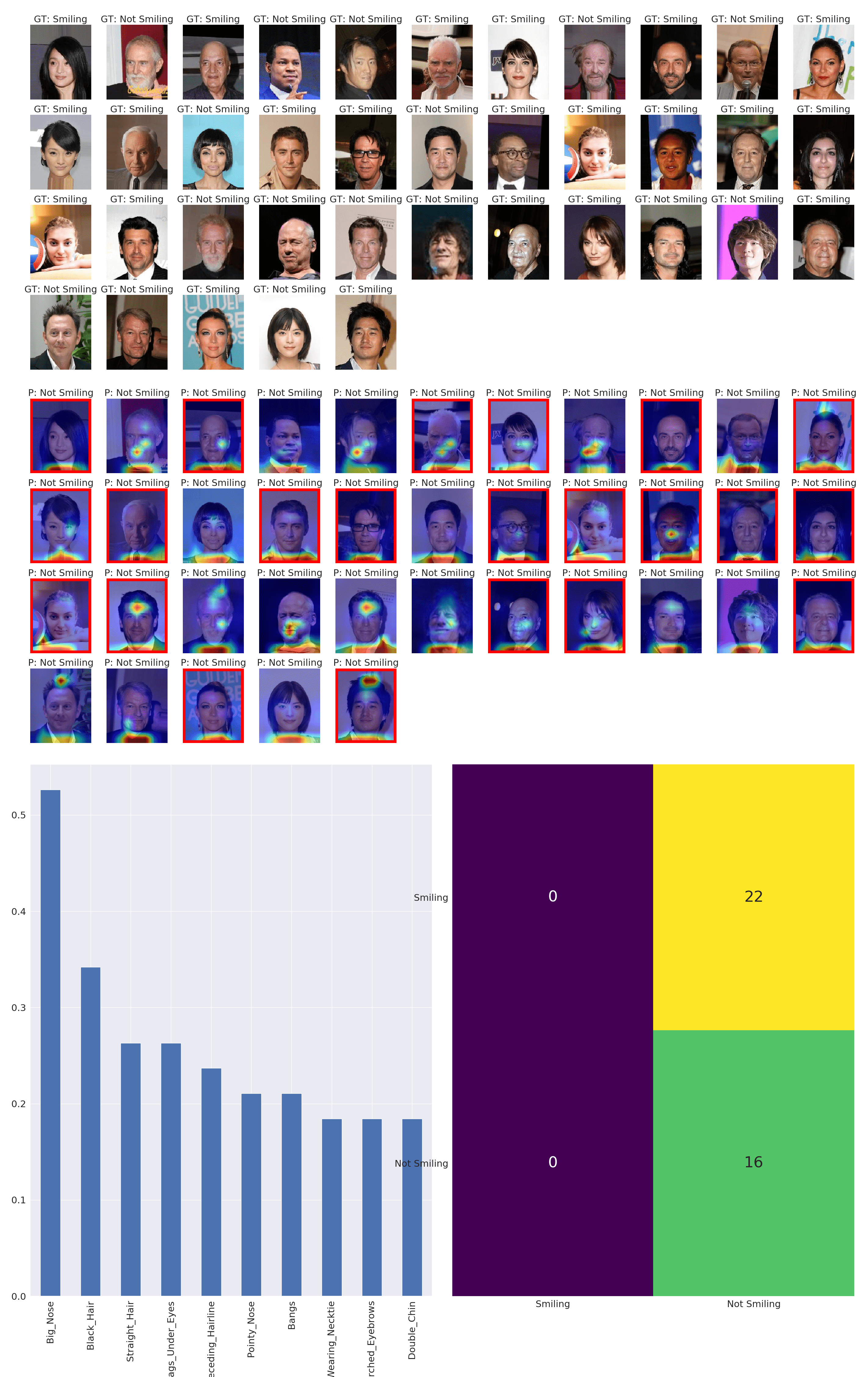}
    \caption{Cluster \#4}
    \label{fig:og-4}
\end{figure}

\begin{figure}[H] 
    \centering
    \includegraphics[width=\textwidth,height=0.9\textheight,keepaspectratio]{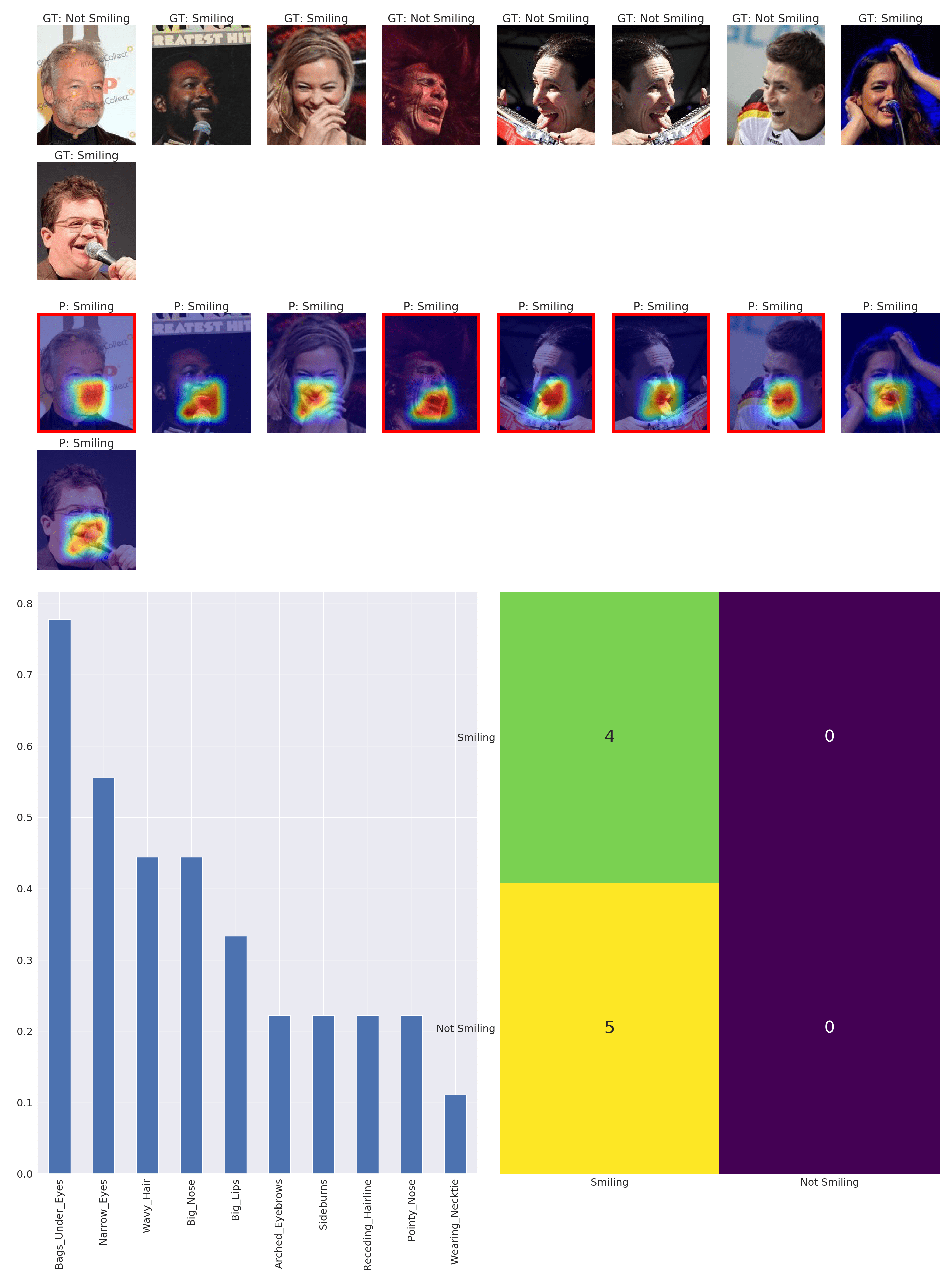}
    \caption{Cluster \#5}
    \label{fig:og-5}
\end{figure}

\begin{figure}[H] 
    \centering
    \includegraphics[width=\textwidth,height=0.9\textheight,keepaspectratio]{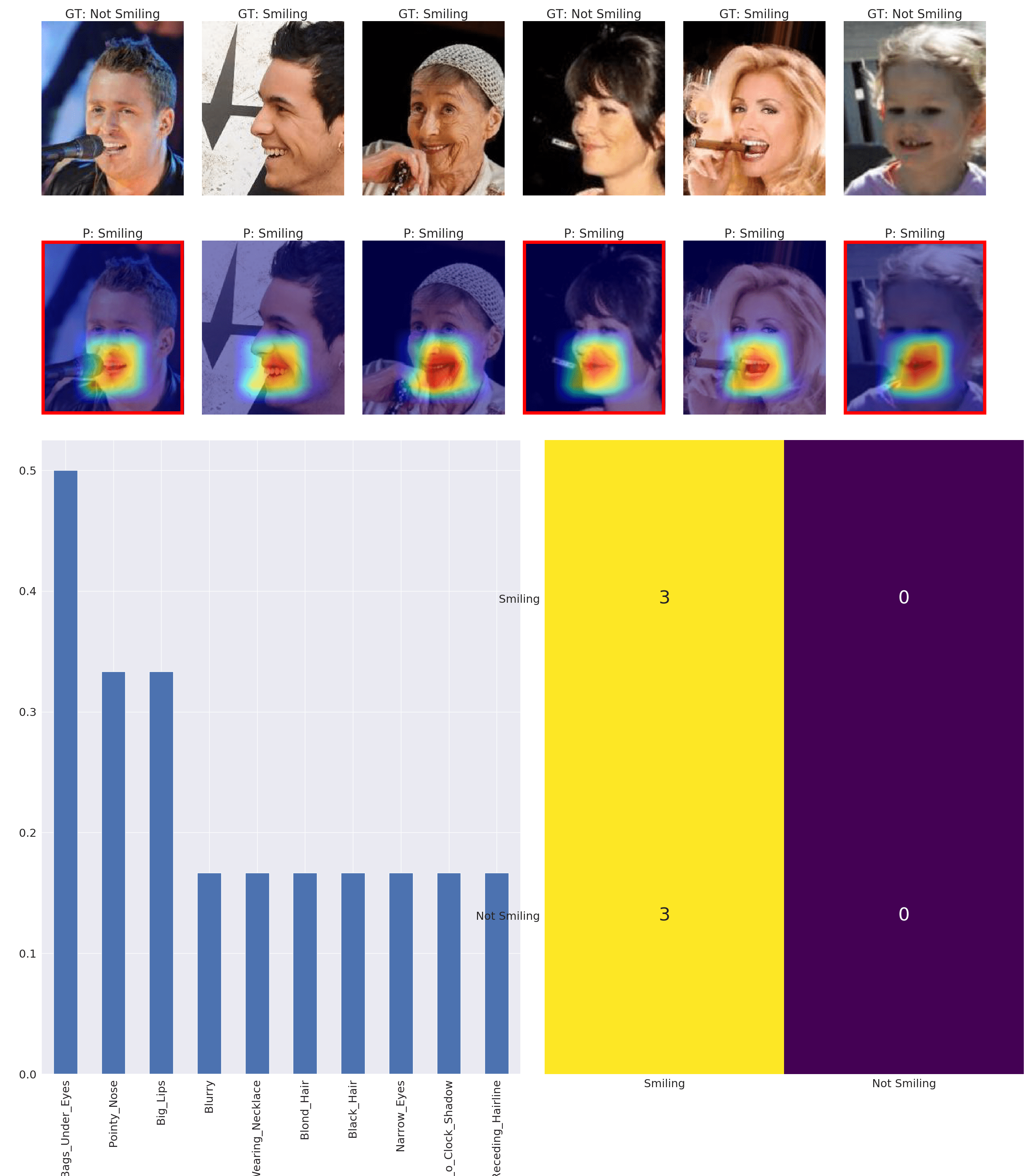}
    \caption{Cluster \#6}
    \label{fig:og-6}
\end{figure}

\begin{figure}[H] 
    \centering
    \includegraphics[width=\textwidth,height=0.9\textheight,keepaspectratio]{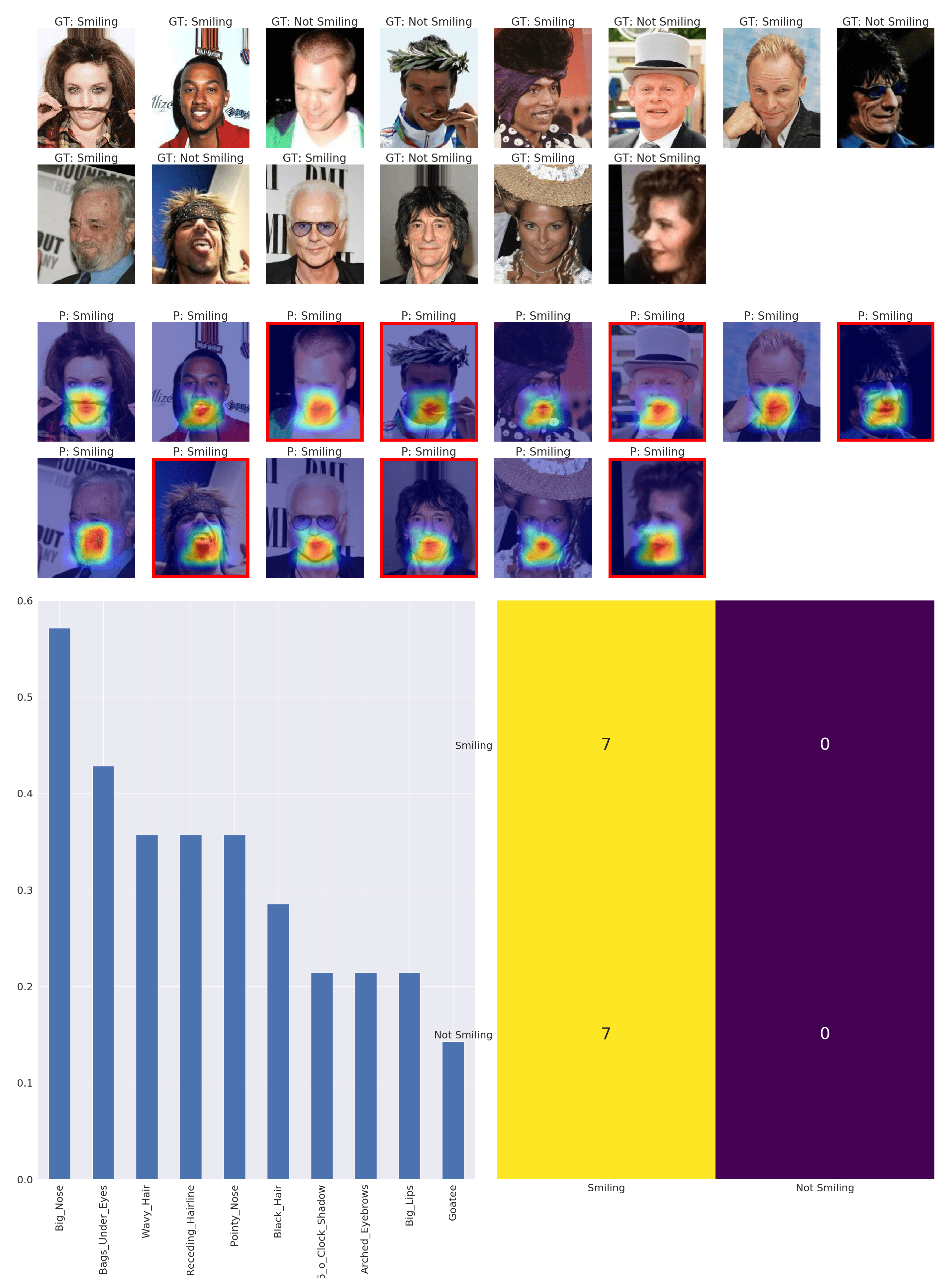}
    \caption{Cluster \#7}
    \label{fig:og-7}
\end{figure}

\begin{figure}[H] 
    \centering
    \includegraphics[width=\textwidth,height=0.9\textheight,keepaspectratio]{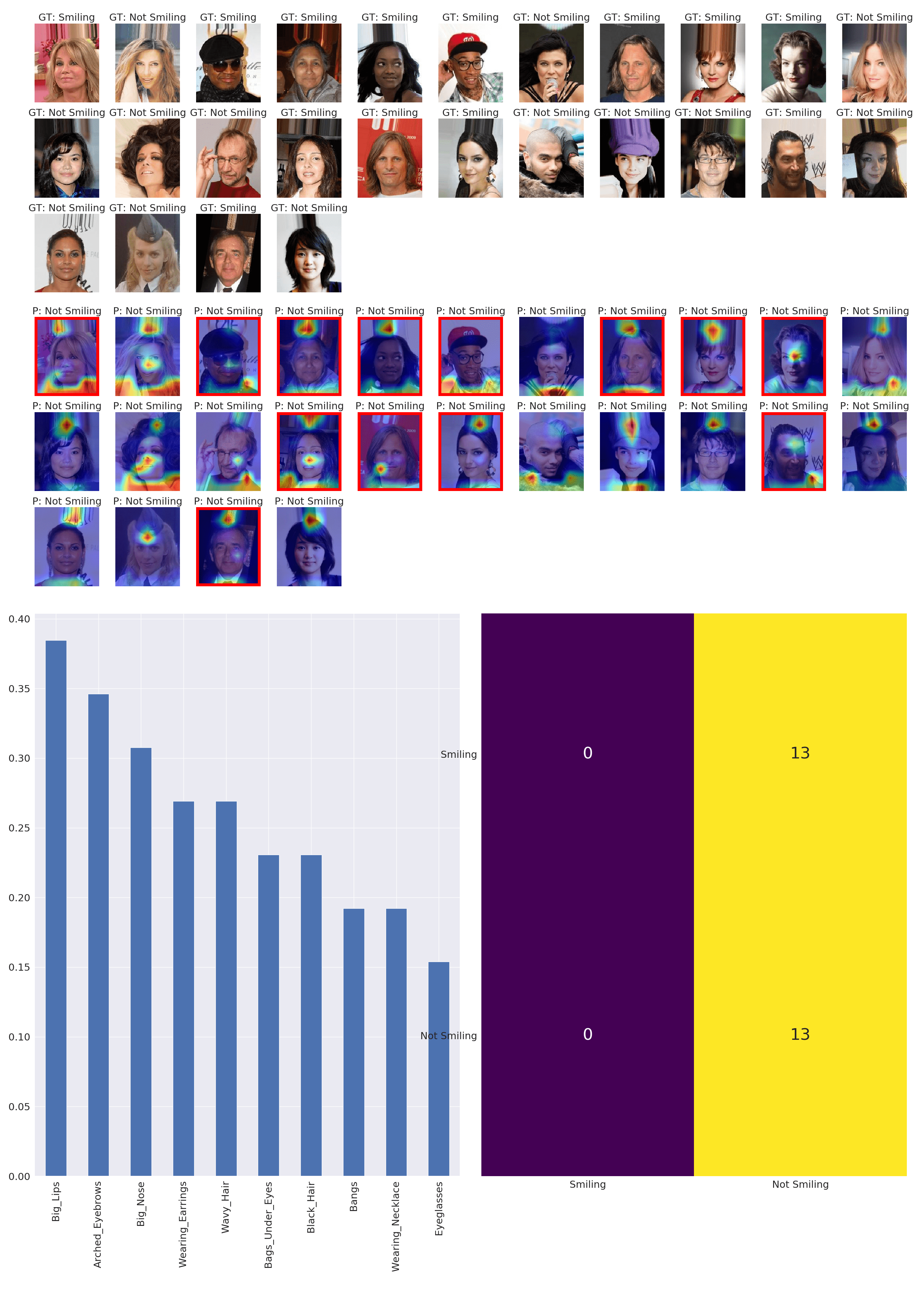}
    \caption{Cluster \#8}
    \label{fig:og-8}
\end{figure}

\begin{figure}[H] 
    \centering
    \includegraphics[width=\textwidth,height=0.9\textheight,keepaspectratio]{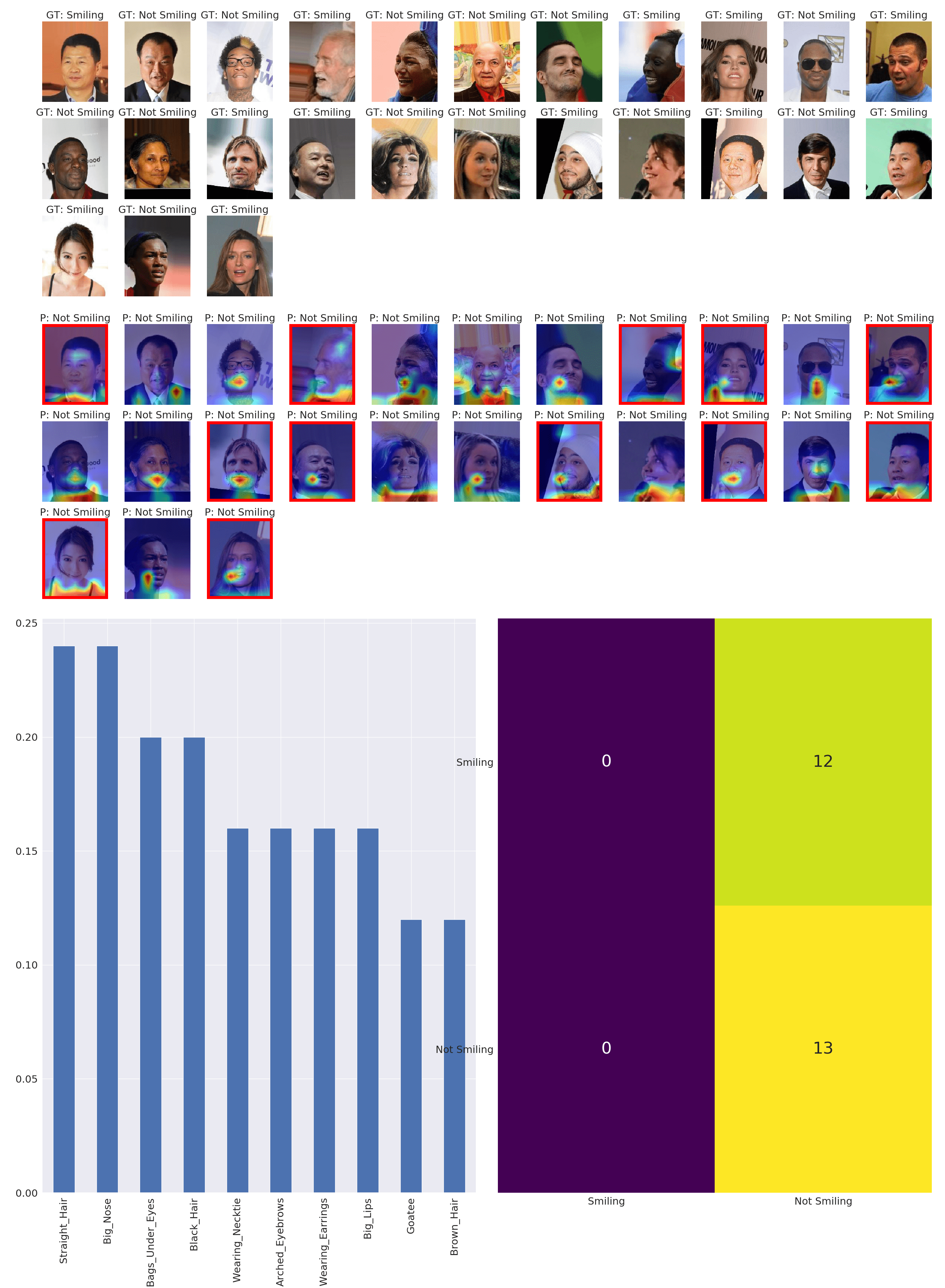}
    \caption{Cluster \#9}
    \label{fig:og-9}
\end{figure}

\begin{figure}[H] 
    \centering
    \includegraphics[width=\textwidth,height=0.9\textheight,keepaspectratio]{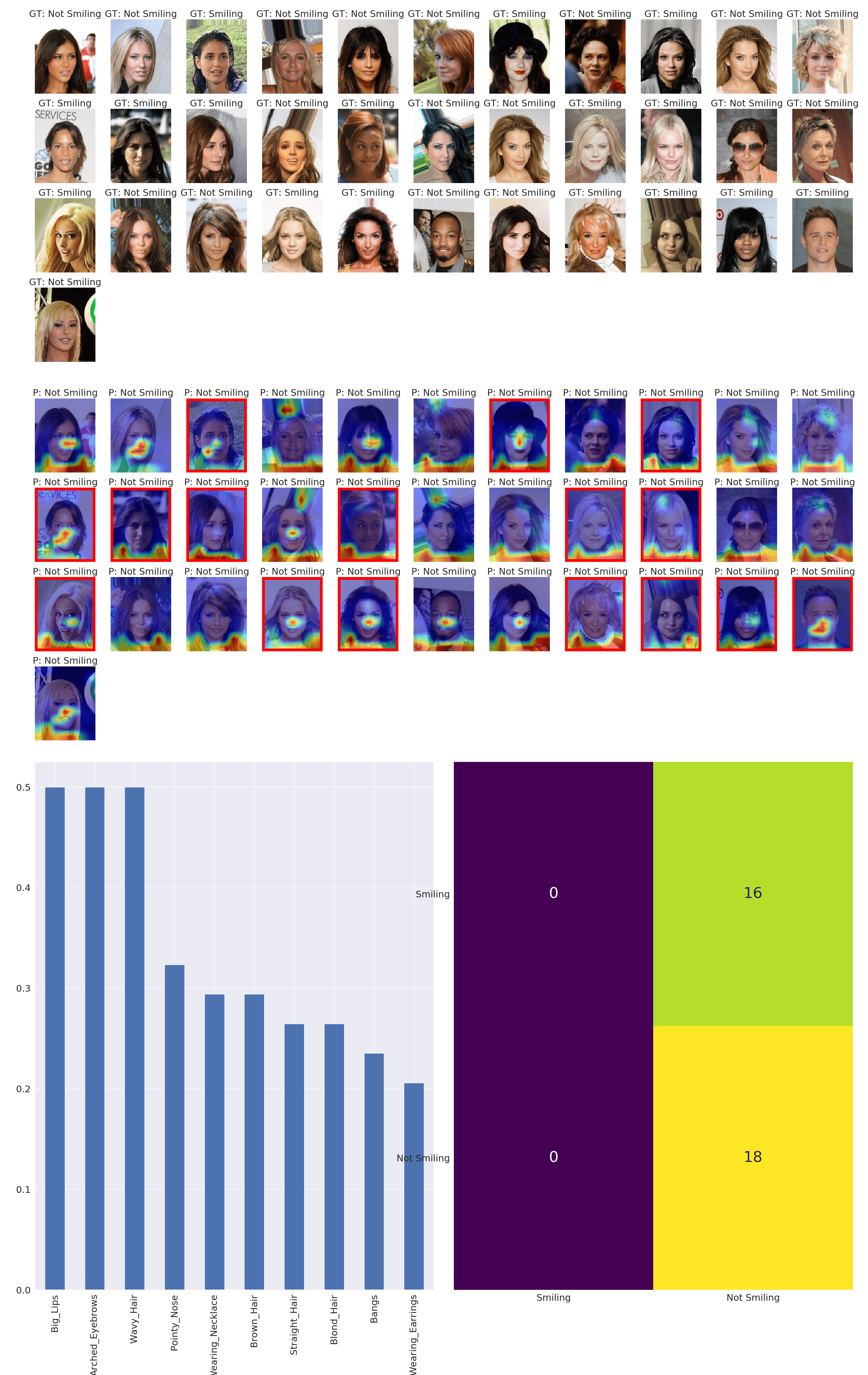}
    \caption{Cluster \#10}
    \label{fig:og-10}
\end{figure}

\subsection{Results on COCO}
\label{appendix:coco}

\subsubsection{Predicted category : \textit{cup}}
\label{appendix:coco-cup}

\begin{figure}[H] 
    \centering
    \includegraphics[width=\textwidth,height=0.9\textheight,keepaspectratio]{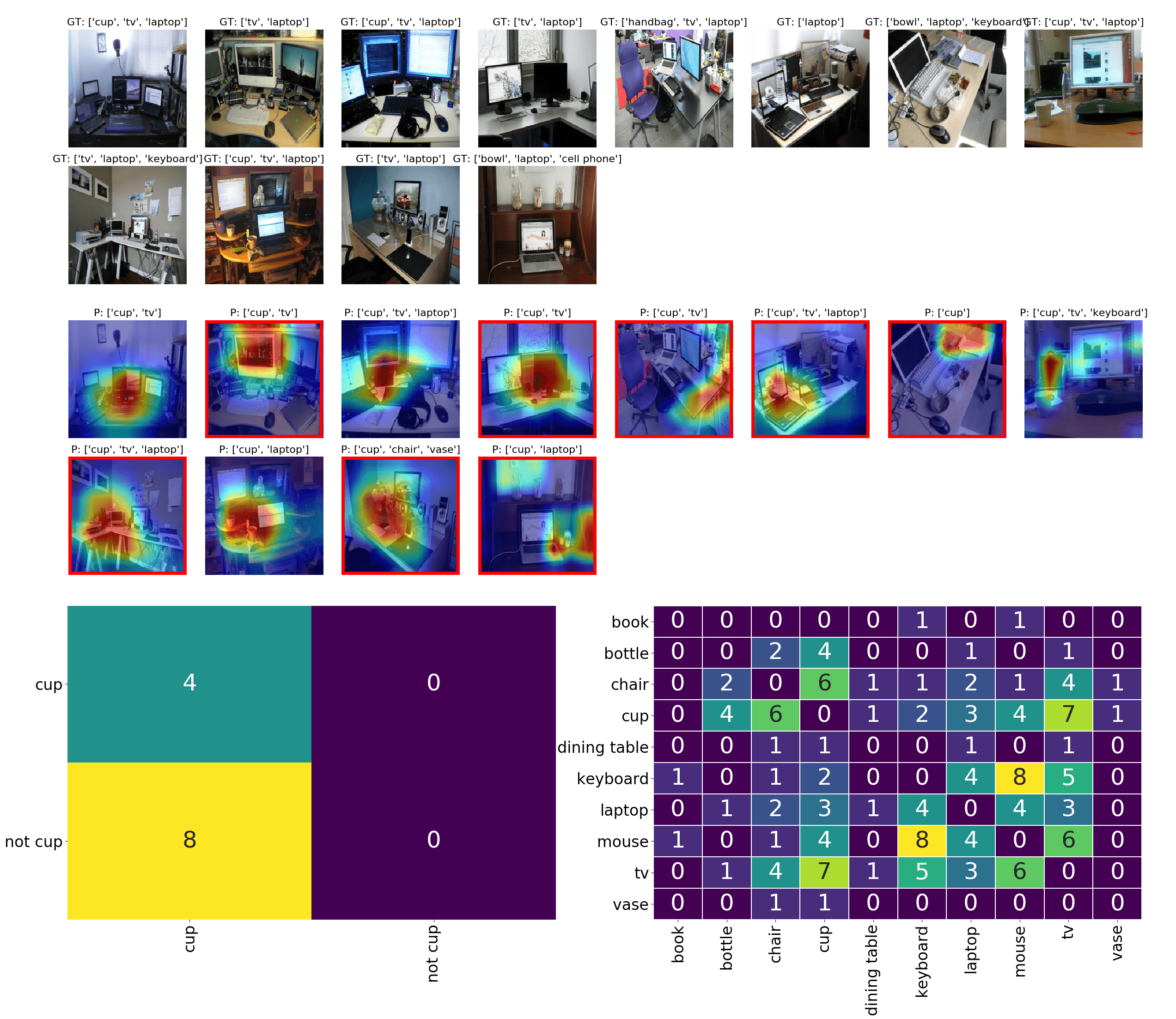}
    \caption{Cluster \#1}
    \label{fig:cup-1}
\end{figure}

\subsubsection{Predicted category : \textit{bed}}
\label{appendix:coco-bed}

\begin{figure}[H] 
    \centering
    \includegraphics[width=\textwidth,height=0.9\textheight,keepaspectratio]{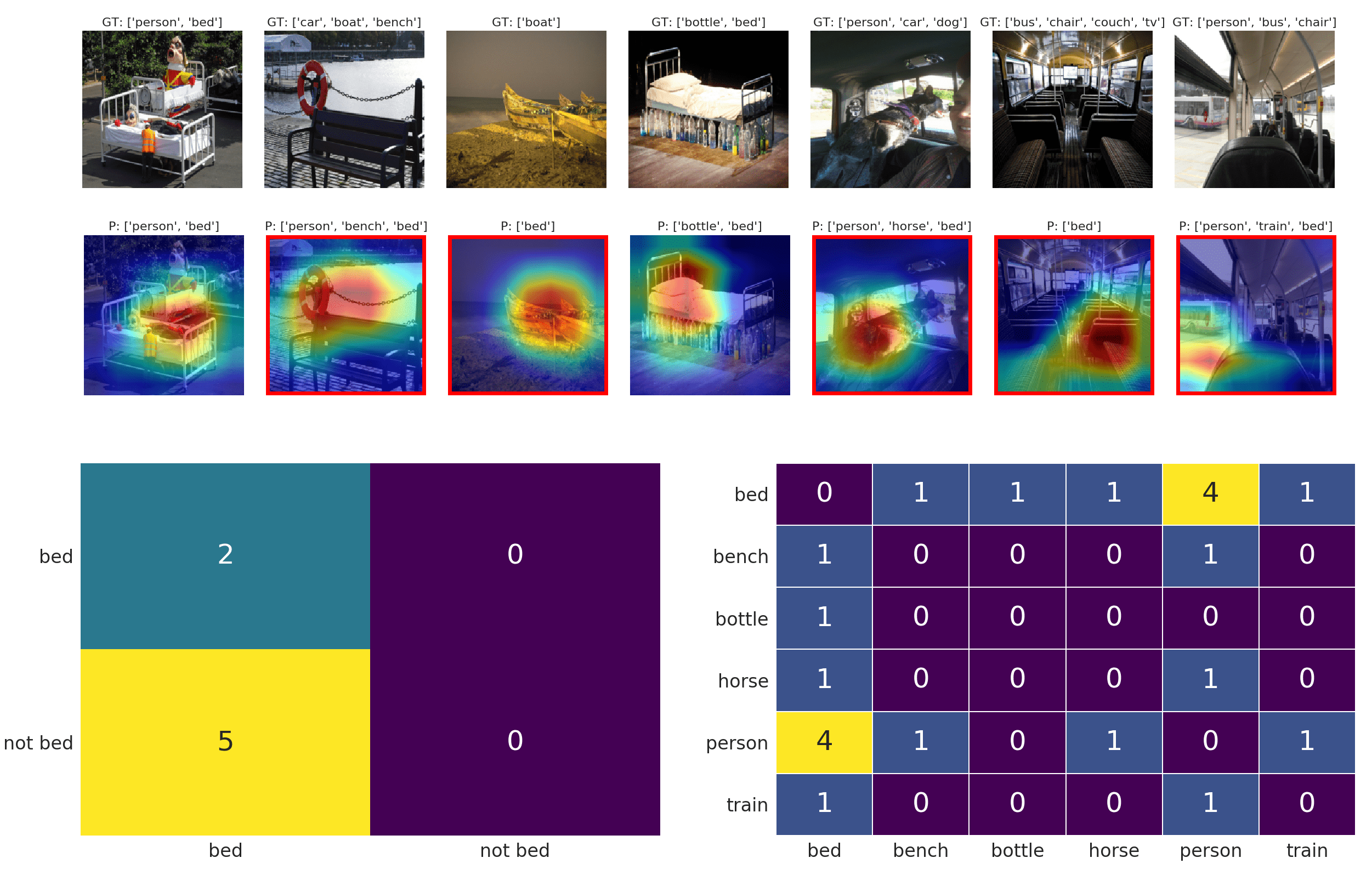}
    \caption{Cluster \#1}
    \label{fig:bed-1}
\end{figure}

\begin{figure}[H] 
    \centering
    \includegraphics[width=\textwidth,height=0.9\textheight,keepaspectratio]{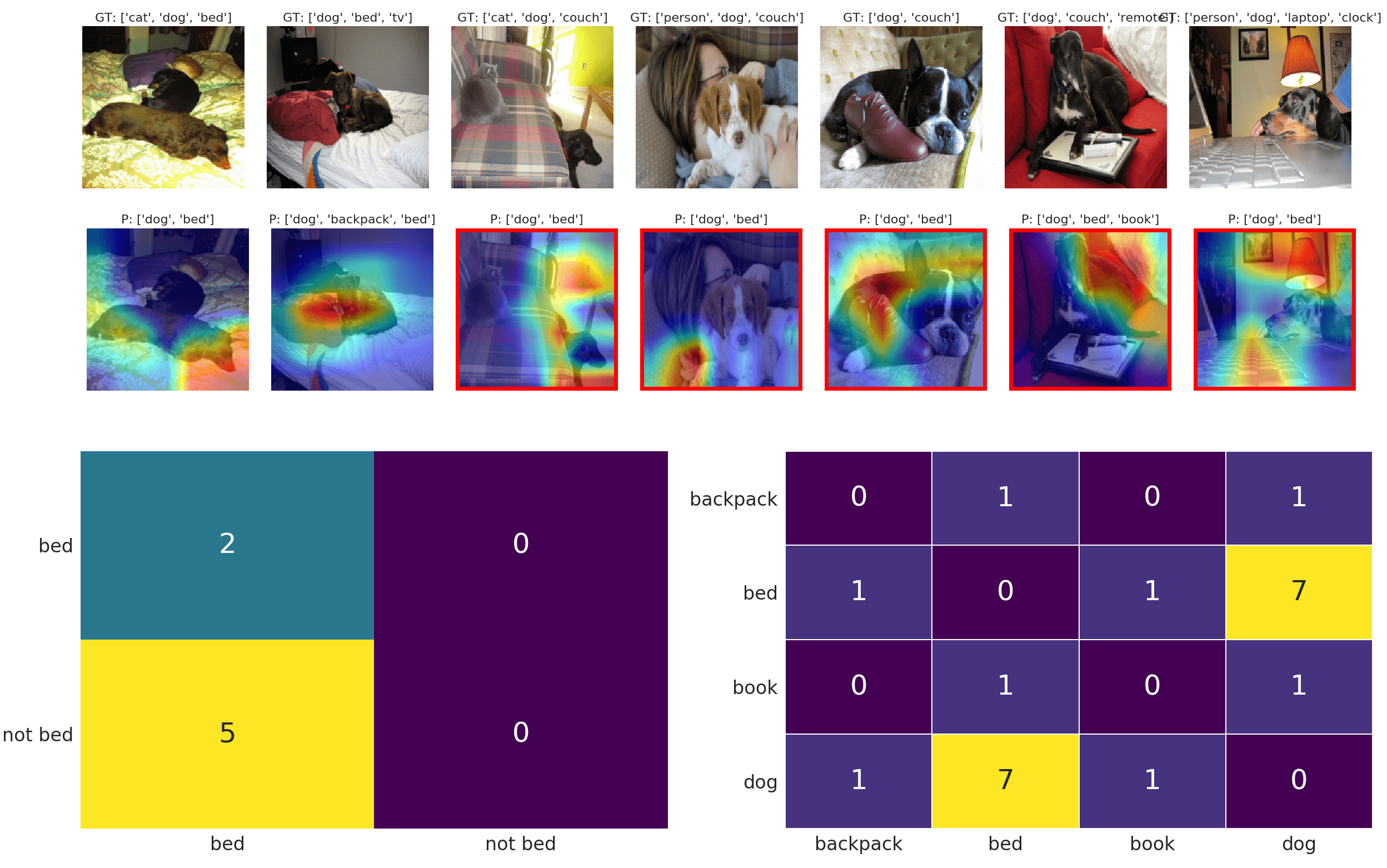}
    \caption{Cluster \#2}
    \label{fig:bed-2}
\end{figure}

\begin{figure}[H] 
    \centering
    \includegraphics[width=\textwidth,height=0.9\textheight,keepaspectratio]{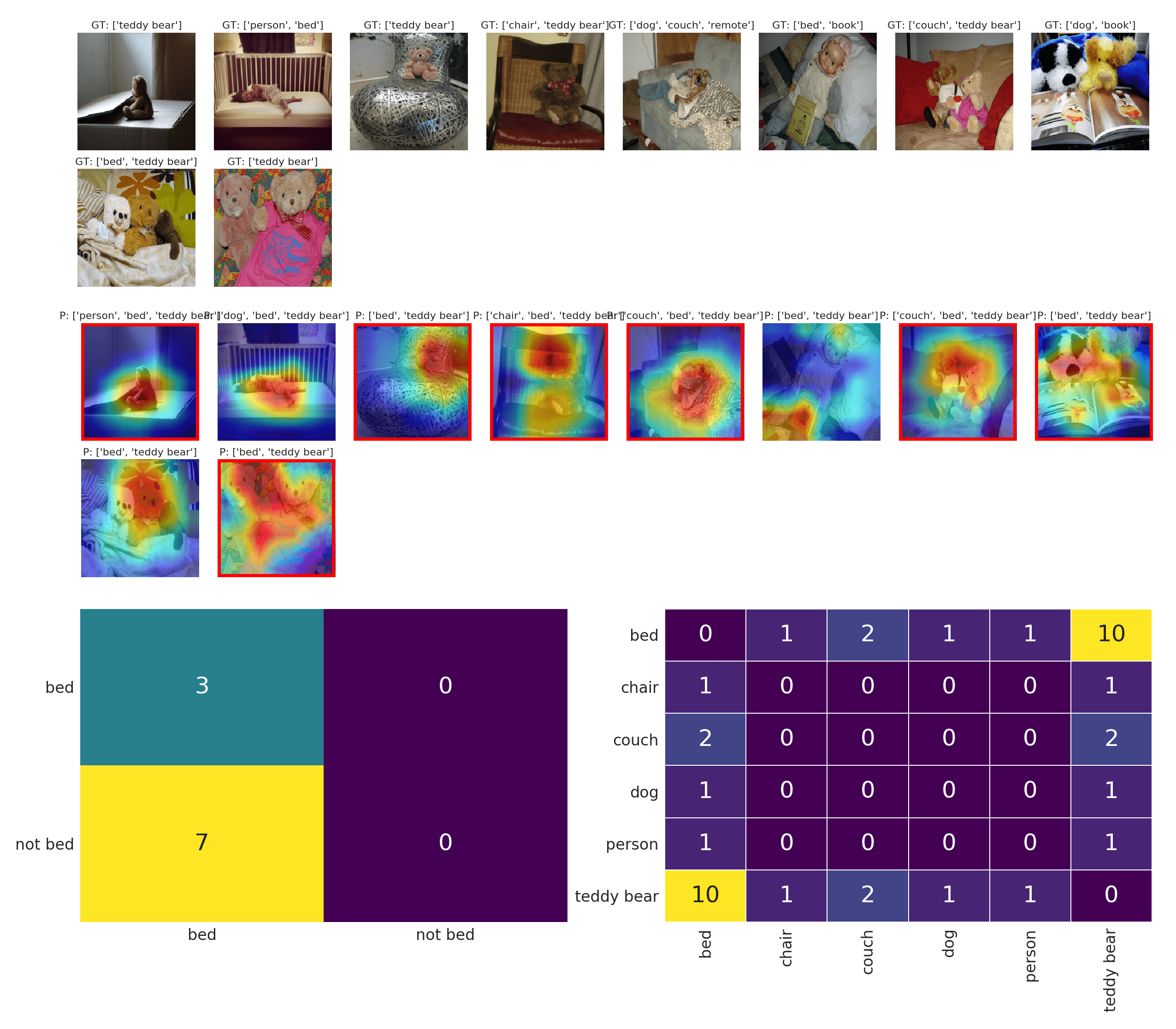}
    \caption{Cluster \#3}
    \label{fig:bed-3}
\end{figure}

\subsubsection{Predicted category : \textit{person}}
\label{appendix:coco-person}

\begin{figure}[H] 
    \centering
    \includegraphics[width=\textwidth,height=0.9\textheight,keepaspectratio]{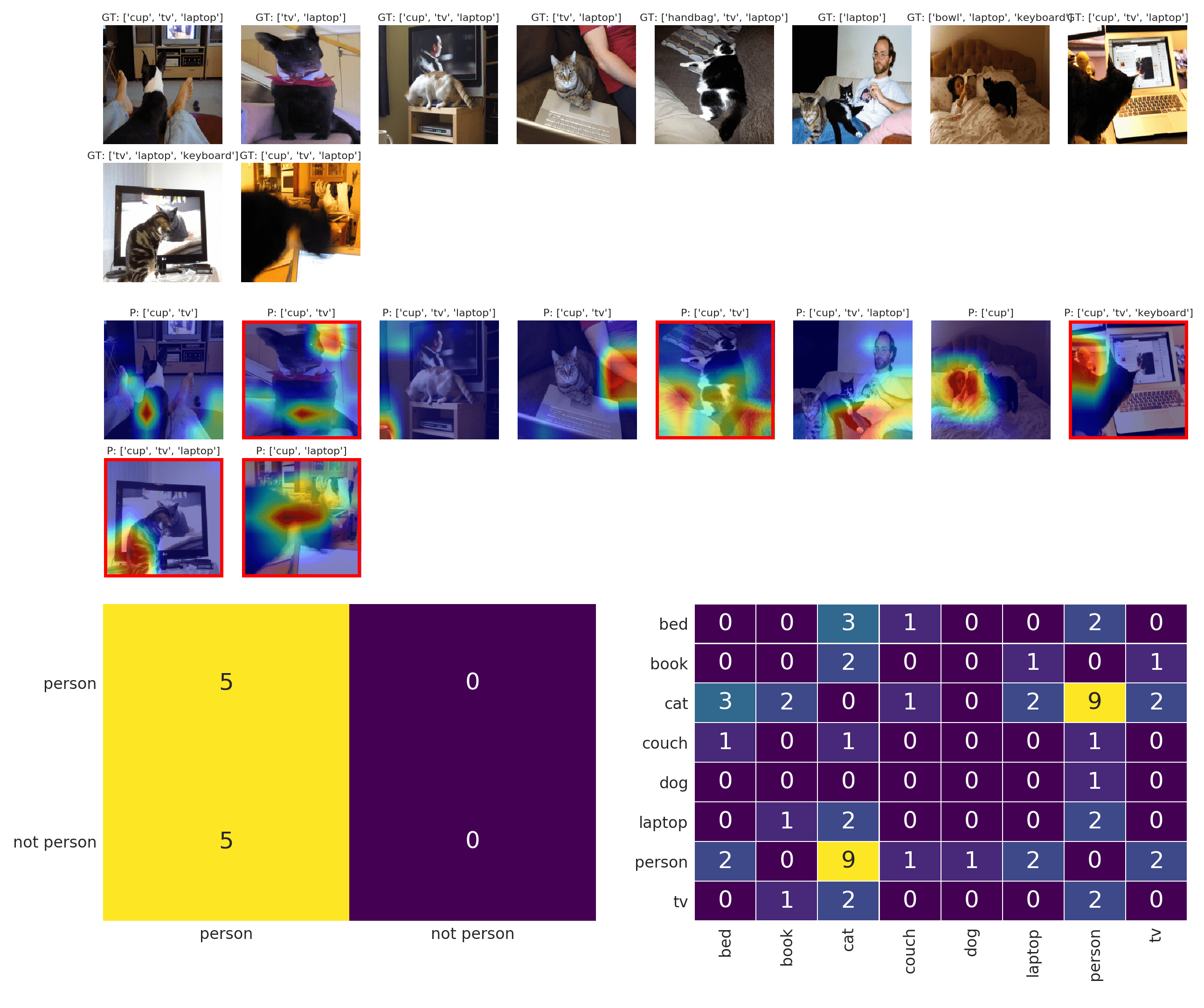}
    \caption{Cluster \#1}
    \label{fig:person-1}
\end{figure} 
\end{document}